\documentclass[dvipsnames,format=sigconf,anonymous=False,review=False,nonacm=True]{acmart}
\AtBeginDocument{%
  }

\setcopyright{acmlicensed}
\copyrightyear{2018}
\acmYear{2018}
\acmDOI{}
\acmConference[The Genetic and Evolutionary Computation Conference]{The Genetic and Evolutionary Computation Conference}{July 14--18,
  2025}{Woodstock, NY}
\acmISBN{978-1-4503-RLDE-AFLX-X/18/06}
\usepackage{multirow}
\usepackage{booktabs}
\usepackage[ruled]{algorithm2e}
\usepackage{subfig}
\usepackage{threeparttable}
\usepackage{booktabs}
\usepackage{nicefrac}
\usepackage{makecell}
\usepackage{longtable}

\begin{document}

\title{Reinforcement Learning-based Self-adaptive Differential Evolution through Automated Landscape Feature Learning }
\renewcommand{\shorttitle}{RL-based DE through Automated Landscape Feature Learning}
\author{Hongshu Guo}
\email{guohongshu369@gmail.com}
\orcid{0000-0001-8063-8984}
\affiliation{%
  \institution{South China University of Technology}
  \city{Guangzhou}
  \state{Guangdong}
  \country{China}
}
\author{Sijie Ma}
\email{masijie9@gmail.com}
\orcid{0009-0008-6129-0191}
\affiliation{%
  \institution{South China University of Technology}
  \city{Guangzhou}
  \state{Guangdong}
  \country{China}
}
\author{Zechuan Huang}
\email{hohqhzc825@gmail.com}
\orcid{0009-0000-3668-4481}
\affiliation{%
  \institution{South China University of Technology}
  \city{Guangzhou}
  \state{Guangdong}
  \country{China}
}
\author{Yuzhi Hu}
\email{goodd0z1y@gmail.com}
\orcid{0009-0009-9960-355X}
\affiliation{%
  \institution{South China University of Technology}
  \city{Guangzhou}
  \state{Guangdong}
  \country{China}
}
\author{Zeyuan Ma}
\email{scut.crazynicolas@gmail.com}
\orcid{0000-0001-6216-9379}
\affiliation{%
  \institution{South China University of Technology}
  \city{Guangzhou}
  \state{Guangdong}
  \country{China}
}
\author{Xinglin Zhang}
\email{csxlzhang@scut.edu.cn}
\orcid{0000-0003-2592-6945}
\affiliation{%
  \institution{South China University of Technology}
  \city{Guangzhou}
  \state{Guangdong}
  \country{China}
}
\author{Yue-Jiao Gong}
\email{gongyuejiao@gmail.com}
\authornote{Corresponding author.}
\orcid{0000-0002-5648-1160}
\affiliation{%
  \institution{South China University of Technology}
  \city{Guangzhou}
  \state{Guangdong}
  \country{China}
}

\renewcommand{\shortauthors}{Guo et al.}

\begin{abstract}
  Recently, Meta-Black-Box-Optimization (MetaBBO) methods significantly enhance the performance of traditional black-box optimizers through meta-learning flexible and generalizable meta-level policies that excel in dynamic algorithm configuration (DAC) tasks within the low-level optimization, reducing the expertise required to adapt optimizers for novel optimization tasks. Though promising, existing MetaBBO methods heavily rely on  human-crafted feature extraction approach to secure learning effectiveness. To address this issue, this paper introduces a novel  MetaBBO method that supports automated feature learning during the meta-learning process, termed as RLDE-AFL, which integrates a learnable feature extraction module into a reinforcement learning-based DE method to learn both the feature encoding and meta-level policy. Specifically,  we design an attention-based neural network with mantissa-exponent based embedding to transform the solution populations and corresponding objective values during the low-level optimization into expressive landscape features. We further incorporate a comprehensive algorithm configuration space including diverse DE operators into a  reinforcement learning-aided DAC paradigm to unleash the behavior diversity and performance of the proposed RLDE-AFL. Extensive benchmark results show that co-training the proposed feature learning module and DAC policy contributes to the superior optimization performance of RLDE-AFL to several advanced DE methods and recent MetaBBO baselines over both synthetic and realistic BBO scenarios. The source codes of RLDE-AFL are available at \url{https://github.com/GMC-DRL/RLDE-AFL}.
\end{abstract}



\maketitle
\section{Introduction}

In the last few decades, Evolutionary Computation (EC) methods such as Genetic Algorithm (GA)~\cite{GA}, Particle Swarm Optimization (PSO)~\cite{PSO} and Differential Evolution (DE)~\cite{DE} have become more and more eye-catching in solving Black-Box Optimization (BBO) problems lacking accessible formulations or derivative information in both academia and industry~\cite{ma2024toward}. However, according to the ``No-Free-Lunch'' (NFL) theorem~\cite{wolpert1995no}, no single algorithm or algorithm configuration (AC) can dominate on all problems. Therefore, to enhance the optimization performance on diverse problems, researchers have manually designed various adaptive operator selection and parameter control methods, which achieve superior performance on BBO benchmarks~\cite{2009bbob,cec2021,metabox}. However, the design of the operator and parameter adaptive mechanisms requires substantial experience and deep expertise in optimization problems and algorithms. One may need to adjust the configurations iteratively according to the problem characteristic and optimization feedback, consuming significant time and computational resources. 

To relieve the human effort burden, recent researchers introduce Meta-Black-Box Optimization (MetaBBO) which leverages a meta-level policy to replace the human-crafted algorithm designs in Algorithm Selection~\cite{RLDAS}, Algorithm Configuration~\cite{GLEET,rlemmo,LDE,DEDDQN}, Solution Manipulation~\cite{li2024glhf,li2024bopt,wu2023decn,li2024pretrained} and Algorithm Generation~\cite{symbol,llamoco,aldes}. MetaBBO methods typically involve a bi-level architecture. In the meta level, given the optimization state, a neural network based policy determines an appropriate algorithm design for the low-level algorithm at each optimization generation. The resulting changes in objective values after algorithm optimization are returned to the meta-level policy as meta performance signal, which is then used to refine the meta policy~\cite{yang4956956meta,ma2024toward}. Given the meta policy $\pi_\theta$ parameterized by $\theta$ with the algorithm $A$ on a set of problem instances $\mathcal{I}$, the objective for algorithm configuration is formulated as:
\begin{equation}\label{eq:metabbo}
    \mathbb{J}(\theta) = \mathop{\arg\max}\limits_{\theta \in \Theta} \mathop{\mathbb{E}}\limits_{f\in \mathcal{I}} \big[ \sum_{t=1}^{T}Perf(A, \omega_t, f) \big]
\end{equation}
where $T$ is the optimization horizon, $Perf(\cdot)$ is a performance metric function, $\omega_t=\pi_\theta(s_t)$ is the algorithm design outputted by the policy and $s_t$ is the optimization state at generation $t$. For algorithm configuration, $\omega$ may correspond to operator selection, parameter values, or both. To maximize the expected performance on the problem distribution, machine learning methods such as Reinforcement Learning (RL) are widely adopted for policy training. 

\begin{figure}[t]
\centering
\includegraphics[width=0.95\columnwidth]{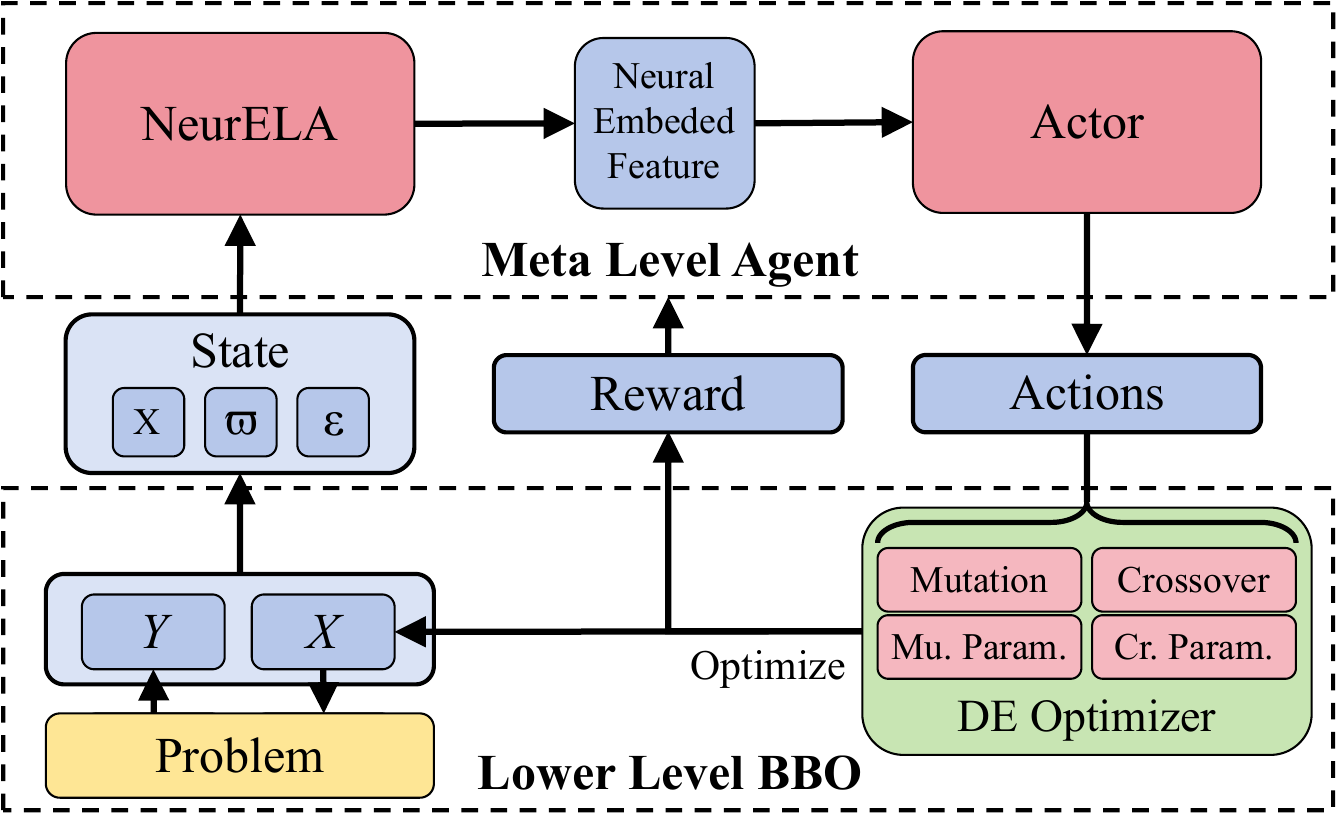}
\caption{The overview of the bi-level structure in RLDE-AFL.}
\label{fig:overview}
\end{figure}

Though promising, these works still retain significant potential to further reduce the expertise burden and enhance the performance. The first limitation is the expertise dependent optimization state design, which requires substantial domain to develop informing and representative features for comprehensive configurations. 
Besides, while traditional BBO community have developed diverse operator variants, existing MetaBBO methods typically adopt only a small subset of these operators, leading to a limited strategy diversity. 
Moreover, the configuration spaces in existing MetaBBO methods are restricted, most works focus on exclusively operator selection~\cite{DEDDQN,DEDQN} or parameter control~\cite{LDE,GLEET}, failing to fully unleash the behavior diversity of the meta policies. 

To address these issues, we propose RLDE-AFL. 
Firstly, for the expertise-dependent state representation, we employ NeurELA~\cite{neurela}, as shown in the top left of Figure~\ref{fig:overview}, a self-attention neural network based optimization information extractor to automatically analyze the problem landscape from the population and evaluation values, eliminating the need for manual feature design. The two-stage self-attention between the dimensions and individuals achieves generalizable and comprehensive state extraction. 
By integrating the mantissa-exponent based evaluation value representation, we further enhance the generalization ability of RLDE-AFL.

Then, to further enhance the performance of RLDE-AFL, we integrate 14 DE mutation operators and 3 DE crossover operators with diverse optimization behaviours to form the candidate operator pool (illustrated in the bottom right of Figure~\ref{fig:overview}). This diversity enables the agent to adaptively deploy distinct optimization strategies tailored to different problem instances. By incorporating parameter control, RL agent gains full control over the DE framework through optimization states derived from the automatic feature extractor, thereby unleashing the behavior diversity of RL agent to achieve superior performance.

Finally, we conduct extensive experiments to demonstrate the effectiveness of RLDE-AFL on MetaBox~\cite{metabox} benchmark problems. The comparisons with advanced traditional DE algorithms and RL-based DE methods confirm the superior performance of RLDE-AFL. The zero-shot performance on different dimensional problems, expensive problems and out-of-distribution realistic problems shows the robust generalization capabilities of RLDE-AFL, outperforming existing advanced traditional and learning based methods. 

The rest of this paper is organized as follows: 
Section~\ref{sec:relatedwork} reviews the related works on existing traditional adaptive DE. 
Section~\ref{sec:perlim} introduces the preliminary concepts on RL and NeurELA. 
In Section~\ref{sec:methodology}, we present our technical details, including the Markov Decision Process (MDP) definition, network design and training process. 
The experimental results are presented in Section~\ref{sec:Exp}, followed by a conclusion in Section~\ref{sec:conclu}.

\section{Related Works}\label{sec:relatedwork}

\subsection{Traditional AC for DE}

For DE algorithms, the algorithm configurations usually focus on selecting mutation and crossover operators and controlling the parameters of these operators. In vanilla DE~\cite{DE}, the used operator and parameters are static across optimization horizon and problem instances. However, since the landscapes and characteristics of BBO problems can vary, static configurations may not be optimal. 

In the last few decades, DE community has proposed diverse mutation and crossover operators with diverse exploratory and exploitative behaviours. Combining these operators using adaptive operator selection mechanisms is a promising approach for superior performance. 
SaDE~\cite{SaDE} assigns each individual mutation operators selected from the two candidate operators following the probabilities calculated from operators' historical performance. 
CoDE~\cite{CoDE} integrates 3 mutation-crossover combinations and selects the best one among the 3 individuals generated by the three combinations as the offspring. 
LADE~\cite{li2023enhancing} assigns different mutation operators for different sub-populations to enhance the population diversity. 

To further enhance the behavior diversity and improve algorithm performance, jDE~\cite{brest2006self} randomly generates scale factor $F$ for mutation and crossover rate $Cr$ for crossover operators, then keeps the parameters that successfully improved individuals. 
One of the advanced variant of jDE, JDE21~\cite{brest2021self} introduces multi-population mechanism and divides the population into exploratory one and exploitative one with different parameter and operator configurations. 
However, random searching parameter configurations is not efficient, JADE~\cite{JADE} samples parameters from normal or Cauchy distribution with the mean updated by the Lehmer mean of the parameter values that successfully improve the individuals in last generation.
Based on JADE, SHADE~\cite{tanabe2013success} uses two memories to record the Lehmer mean of $F$ and $Cr$ respectively in each generation, which enhances the exploration.
Such successful history based parameter adaptive mechanism is widely adopted in advanced DE variants such as MadDE~\cite{biswas2021improving}, NL-SHADE-LBC~\cite{stanovov2022nl} and L-SRTDE~\cite{L-SRTDE}.

Although these mechanisms bring significant performance~\cite{cec2021,cec2022,cec2024}, one must equip with enough expert knowledge on the algorithm and optimization problem to design suitable operator selection rules and parameter adaptive mechanisms, which leads to heavy human-effort burden. 

\subsection{MetaBBO-RL based AC for DE}

To relieve the burden, researchers turns to machine learning for answers. One of the most commonly adopted solution is to use Reinforcement Learning (RL) agents to learn the knowledge about optimization and dynamically determine the operator selection and parameter control according to the optimization states. 
For operator selection, value-based RL methods such as Tabular Q-Learning~\cite{1992qlearning} and Deep Q-Learning~\cite{2013dqn} are widely adopted to handle the discrete action spaces. Tabular Q-Learning maintains a table mapping each discrete state to an optimal action~\cite{DE-RLFR,MARLwCMA,LRMODE,RL-CORCO,RL-SHADE,RLMMDE,RLEA-SSC,QLSHADE}. For instance, RLDMDE~\cite{RLDMDE} selects mutation-crossover operator combinations for each sub-population according to the population diversity levels. 
However, the coarse-grained discrete state spaces may lead to information loss and degrade the performance. Therefore, neural networks are introduced to process continuous states and predict the expected accumulated reward of each candidate action, which turns to Deep Q-Network (DQN)~\cite{QLSHADE,DEDDQN,DEDQN,HF}. DE-DDQN~\cite{DEDDQN} designs a 99-dimensional continuous state including the optimization status and operator performance history. A Multilayer Perceptron (MLP) based Double DQN~\cite{2016ddqn} agent is employed to determine the mutation operator for each individual. Though promising, the complex state design significantly slows down the time efficiency. DE-DQN~\cite{DEDQN} uses Fitness Landscape Analysis (FLA)~\cite{fla} to extract 4 optimization features from random walk sampling, which are then feed into a DQN agent for operator selection. 
Besides, neural networks can also be used to predict the probabilities of action selection~\cite{rlemmo,PG-DE,SA-DQN-DE,CEDE-DRL}. RLEMMO~\cite{rlemmo} uses the agent trained by Proximal Policy Optimization (PPO)~\cite{ppo} to select DE mutation operators for solving multi-modal problems.
Furthermore, some methods consider the whole DE algorithms as switchable components. RL-DAS~\cite{RLDAS} selects algorithms periodically from three advanced candidate DE algorithms according to the optimization status and algorithm histories.

For parameter control, some methods discretize the continuous action space and use Q-Learning to select the parameter values~\cite{RLDE,qlDE,RLMODE}. RLMODE~\cite{RLMODE} uses the feasible and domination relationship of individuals as the states to control increasing or decreasing the values of scale factors and crossover rates for solving constrained multi-objective problems.
More RL-based parameter control methods for DE use neural networks to predict the distribution of the target parameters~\cite{ada-smoDE,GLEET,RLHDE,LADE,LDE,MTDE-L2T}. LDE~\cite{LDE} leverages Long Short-Term Memory (LSTM) network to sequentially determine the values of scale factors and crossover rates for each individual according to the population fitness histograms. GLEET~\cite{GLEET} further proposes a Transformer-based architecture~\cite{transformer} to balance the exploration-exploitation in DE and PSO by using parameter control.
There are also works that control both operator and parameters~\cite{RL-HPSDE}. RL-HPSDE~\cite{RL-HPSDE} employs Q-table to select the combinations of operator and parameter sampling methods. 

Although these RL based adaptive methods achieve remarkable optimization performance, they still face three limitations. Firstly, the design of state features still require expertise on optimization problems and algorithms so that features can be properly selected to reflect the optimization status. Besides, some features need extra function evaluations for random sampling which occupy the resource for optimization. Secondly, the collected candidate operators in the operator pools of existing methods are limited, only a small number of operators are considered, which limits the strategy diversity and generalization. Finally, most of these methods focus on selecting operators only or controlling parameters only, the learning and generalization ability of RL agents are not fully developed. To address these limitations, we hence in this paper propose RLDE-AFL, which introduces automatic learning based state representation and integrates diverse operators fully controlled by the RL agent for superior performance.

\section{Preliminary}\label{sec:perlim}

\subsection{Markov Decision Process}\label{sec:pre-mdp}

A MDP could be denoted as $\mathcal{M} := <\mathcal{S}, \mathcal{A}, \mathcal{T}, R>$. Given a state $s_t \in \mathcal{S}$ at time step $t$, the policy $\pi$ accordingly determines an action $a_t \in \mathcal{A}$ which interacts with the environment. The next state $s_{t+1}$ is produced by the changed environment through the environment dynamic $\mathcal{T}(s_{t+1} | s_t, a_t)$. A reward function $R: \mathcal{S} \times \mathcal{A} \rightarrow \mathbb{R}$ acts as a performance metric measuring the performance of the actions. 
Those transitions between states and actions achieve a trajectory $\tau := (s_0, a_0, s_1, \cdots, s_T)$. The target of MDP is to find an optimal policy $\pi^*$ that maximizes the accumulated rewards in trajectories:
\begin{equation}
    \pi^* = \mathop{\arg\max}\limits_{\pi \in \Pi}\sum_{t=1}^T \gamma^{t-1}R(s_t, a_t)
\end{equation}
where $\Pi: \mathcal{S} \rightarrow \mathcal{A}$ selects an action with a given state, $\gamma$ is a discount factor and $T$ is the length of trajectory. 

\subsection{Neural Evolutionary Landscape Analysis}\label{sec:pre-nela}

To obtain the optimization status or the problem characteristic, various landscape analysis methods such as Evolutionary Landscape Analysis (ELA) have emerged. Although these approaches provide a comprehensive understanding, they require a certain level of expert knowledge and consume non-negligible computational resource. To address these issues, learning-based ELA methods are proposed to use neural networks to analyze landscapes~\cite{prager2021towards,Deep-ela}. However, these methods are constrained by problem dimensions or profile problems statically, which makes them not applicable for the feature extraction in MetaBBO. Recently, 
Ma et al.~\cite{neurela} proposes Neural Exploratory Landscape Analysis (NeurELA), which employs a two-stage attention-based neural network as a feature extractor. Given a population $X \in \mathbb{R}^{N\times D}$ with $N$ $D$-dimensional solutions and its corresponding evaluation values $Y\in\mathbb{R}^{N}$, the observations $o$ are organized as per-dimensional tuples $\{\{(X_{i,j}, Y_i)\}_{i=1}^N\}_{j=1}^D$, with a shape of $D \times N \times 2$. Then $o$ is embedded by a linear layer and advanced the two-stage attention for information sharing in cross-solution and cross-dimension levels, respectively. The cross-solution attention uses an attention block to enable same dimensions shared by different candidate solutions within the population to exchange information, while the cross-dimension attention further promotes the sharing of information across different dimensions within each candidate. In this way, NeurELA efficiently extracts comprehensive optimization state information for each individual automatically. The attention-based architectures on both dimension and solution levels boost the scalability of NeurELA across different problem dimensions and different algorithm population sizes.

\section{Methodology}\label{sec:methodology}

\subsection{MDP Formulation}

At the $t$-th generation, given a state $s_t = \{X_t, Y_t, t\}$ including the population $X_t\in \mathbb{R}^{N\times D}$ with $N$ $D$-dimensional individuals and the corresponding evaluation values $Y_t = f(X_t)$ under problem instance $f$, in the meta level the RLDE-AFL policy $\pi_\theta$ parameterized by $\theta$ extracts the optimization features of all individuals using the modified NeurELA module from $s_t$, then determines the operator selection and parameter control actions $a_t \sim \pi_\theta(s_t)$ for each individual. With $a_t$ as configurations, the DE algorithm in the lower level optimize the population and produces the next state $s_{t+1} = \{X_{t+1}, Y_{t+1}, t+1\}$. The reward function $R$ is introduced to evaluate the performance improvement $r_t = R(s_t, a_t|f)$. Considering a set of problem instances $\mathcal{I}$, RL agent targets at finding an optimal policy $\pi_{\theta^*}$ which maximizes the expected accumulated rewards over all problem instances $f\in\mathcal{I}$:
\begin{equation}
    \theta^* = \mathop{\arg\max}\limits_{\theta\in\Theta} \mathop{\mathbb{E}}\limits_{f\in\mathcal{I}}\left[ \sum_{t=1}^{T}\gamma^{t-1}R(s_t, \pi_\theta(s_t)|f)\right]
\end{equation}
In this paper, we use the Proximal Policy Optimization (PPO)~\cite{ppo} to train the policy. 
Next we introduce the detailed MDP designs, including the state, action and reward in the following subsections.

\subsubsection{State.}\label{sec:state}

As mentioned above, the state comprises the population solutions $X_t = \{x_{t,i}\}_{i=1}^N$ and evaluation values $Y_t \in \mathbb{R}^N$ which indicates the optimization situation, and the time stamp $t \in [1, T]$ indicating the optimization progress. Since the searching spaces of different optimization problems vary, we normalize the solution values with the upper and lower bounds of the searching space: $x'_{t,i} = \{\frac{x_{t,i,j}}{ub_j - lb_j}\}_{j=1}^D$ where $ub_j$ and $lb_j$ are the upper and lower bounds of the searching space at the $j$-th dimension, respectively. Besides, the objective value scales across different problem instances can also vary, to ensure state values across problem instances share the same numerical level, we introduce the mantissa-exponent representation for the evaluation value terms in states. Specifically, for each evaluation value $y_{t,i} \in Y_t$, we first represent it in scientific notation $y_{t,i} = \varpi\times10^{e}$, $\varpi\in[-1, 1]$, $e\in\mathbb{Z}$. Then we use a tuple $\{\varpi_{t,i}, \epsilon_{t,i}\}$ as the mantissa-exponent representation of $y_{t,i}$ in the state, where $\epsilon = \frac{e}{\eta}$ and $\eta$ is a scale factor making the scales of exponents in all problem instances similar. For the time stamp $t$ in the state, we normalize it with the optimization horizon $T$ so that it would share the same scale with other features: $s_{time} = \frac{t}{T}$. In summary, the state at generation $t$ is represented as $s_t = \{\{\{\frac{x_{t,i,j}}{ub_j - lb_j}\}_{j=1}^D\}_{i=1}^N, \{(\varpi_{t,i}, \epsilon_{t,i})\}_{i=1}^N, s_{time}\}$, and the observation $o_t$ for NeurELA module is changed to $\{\{(x_{t,i,j}, \varpi_{t,i}, \epsilon_{t,i})\}_{i=1}^N\}_{j=1}^D$.

\subsubsection{Action.}\label{sec:action}

For operator selection we integrate 14 mutation operators and 3 crossover operators form various DE variants.  
\begin{enumerate}
    \item Mutation operators 1$\sim$7 are basic mutation operators in vanilla DE~\cite{DE}: \textit{rand/1}, \textit{best/1}, \textit{rand/2}, \textit{best/2}, \textit{current-to-rand/1}. \textit{current-to-best/1} and \textit{rand-to-best/1}, which have diverse preferences on exploration or exploitation. 
    \item Mutation operator 8$\sim$11 are the variants of the basic operators: \textit{current-to-pbest/1}, \textit{current-to-pbest/1 + archive}, \textit{current-to-rand/1 + archive} and \textit{weighted-rand-to-pbest/1}. They introduce the ``pbest'' technique which replace the best individual by randomly selected top individuals and employ the archive of the eliminated individuals to enhance the exploration.
    \item Mutation operator 12 \textit{ProDE-rand/1}~\cite{PRO-DE} is a variant of \textit{rand/1} which selects random individuals according to the probabilities inversely proportional to distances between individuals.
    \item Mutation operator 13 \textit{HARDDE-current-to-pbest/2}~\cite{harde} improves the optimization performance by introducing time stamps to the archive in \textit{current-to-pbest/1} and sampling random individuals from the two archives containing recent individuals and former individuals respectively.
    \item Mutation operator 14 \textit{TopoDE-rand/1}~\cite{TopoMut-DE} enhances the exploitation of \textit{rand/1} by using the nearest best individual to the current individual as $x_{r1}$.
    \item Crossover operator 1$\sim$2 are basic crossover operators proposed in vanilla DE~\cite{DE}: \textit{binomial} and \textit{exponential} crossover.
    \item Crossover operator 3 \textit{p-binomial} crossover is a variant of \textit{binomial} crossover which borrows the idea of ``pbest'' and replace the parent individual in the crossover with a randomly selected top individual.
\end{enumerate}
In the action space for each individual we include two actions for operator selection  $a^{os1}\in[1,14]$ and $a^{os2}\in[1,3]$. 
The detailed introduction of the operators are shown in Table I in Appendix A.

Besides, fine-tuning the parameters of the selected operators is also a key to superior performance. Therefore, the action space for each individual in this paper includes two parameter control actions $a^{pc1}\in[0,1]^{M_1}$ and $a^{pc2}\in[0,1]^{M_2}$for the selected mutation operators and crossover operators respectively. $M_1$ and $M_2$ are the maximal numbers of parameters of all integrated mutation and crossover operators respectively. In summary, the overall action space for the population at generation $t$ is $a_t = \{(a^{os1}_{i,t}, a^{pc1}_{i,t}, a^{os2}_{i,t}, a^{pc2}_{i,t})\}_{i=1}^{N}$.

\begin{figure*}[t]
\centering
\includegraphics[width=0.85\textwidth]{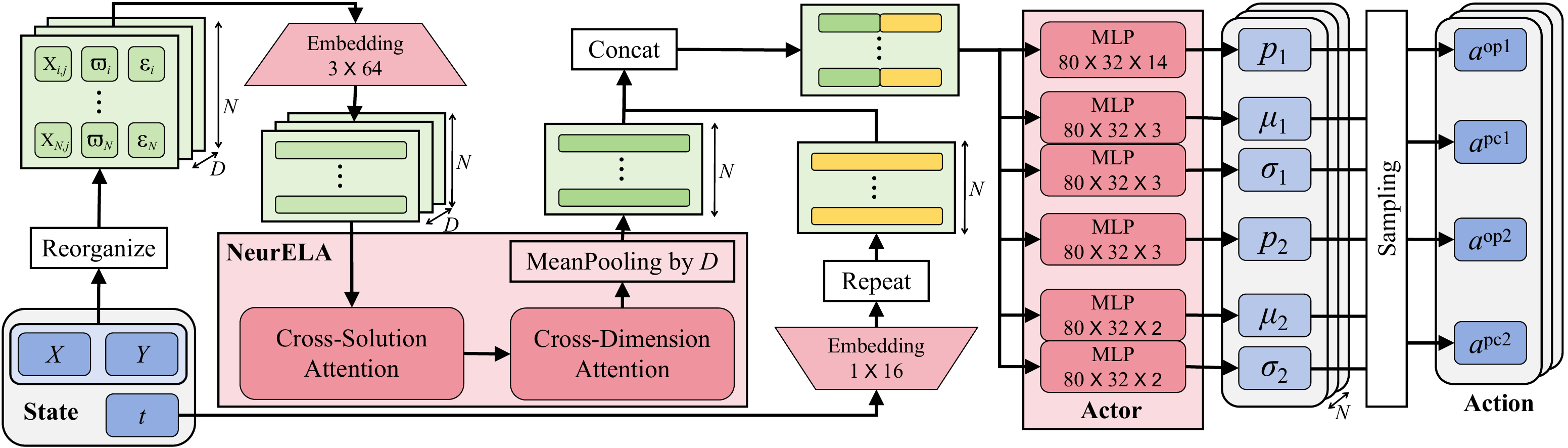}
\caption{Illustration of the network structure.}
\label{fig:network}
\end{figure*}

\subsubsection{Reward.}\label{sec:reward}
We use the reward function formulated as follow:
\begin{equation}\label{eq:reward}
    r_t = \frac{y^*_{t-1} - y^*_{t}}{y^*_0 - y^*}
\end{equation}
where $y^*_t$ is the found best evaluation at generation $t$, $y^*_0$ is the best value in the initial population and $y^*$ is the global optimal evaluation value of the problem instance. The accumulated rewards reflects the optimization performance, and the denominator normalizes all reward values into [0, 1] to make the scales of the accumulated rewards in all problems similar, hence stabilize the training. 

\subsection{Network Design}\label{sec:network}

\subsubsection{Embedding.}
As shown in Figure~\ref{fig:network}, given the state $s$, we first reorganize the population $X$ and evaluation values $Y$ into the observation $o\in\mathbb{R}^{D\times N\times3}$ as mentioned in Section~\ref{sec:state}. Then $o$ is embedded by a linear embedder $\textbf{h}^{(0)} = \phi(o, \textbf{W}_e^{ob})$ where $\phi(\cdot, \textbf{W}_e^{ob})$ denotes a MLP layer with shape $3\times64$. 

\subsubsection{NeurELA.}
The hidden representation $\textbf{h}^{(0)}\in\mathbb{R}^{D\times N\times64}$ is next encoded by the two-stage NeurELA module. In the cross-solution attention, information are shared between the representations of all solutions at the same dimension:
\begin{equation}
\begin{split}
    \hat{\textbf{h}}^{(1)} = & \text{LN}(MSA_S(\textbf{h}^{(0)}) + \textbf{h}^{(0)})\\
    \textbf{h}^{(1)} = & \text{LN}(\phi(\hat{\textbf{h}}^{(1)}; \textbf{W}^{(1)}_S)) + \hat{\textbf{h}}^{(1)})
\end{split}
\end{equation}
where LN denotes the Layernorm~\cite{ba2016layer}, $MSA_S$ is the Multi-head Self-Attention~\cite{transformer} in the solution dimension and $\phi(\cdot;\textbf{W}^{(1)}_S)$ is a MLP layer with the shape of $64\times64$. In the cross-dimension attention, the encoded results $\textbf{h}^{(1)}$ from the cross-solution attention is transposed to $N\times D\times64$ and augmented with cosine/sine positional encodings to maintain the dimensional order within a solution.
\begin{equation}
    \textbf{h}'^{(1)} = (\textbf{h}^{(1)})^\text{T} + \textbf{W}_{pos}
\end{equation}
where $\textbf{W}_{pos}$ is the positional encoding weights. 
Subsequently, we conduct the feature extraction to discover the inner connections between the dimensions:
\begin{equation}
\begin{split}
    \hat{\textbf{h}}^{(2)} = & \text{LN}(MSA_D(\textbf{h}'^{(1)}) + \textbf{h}'^{(1)})\\
    \textbf{h}^{(2)} = & \text{LN}(\phi(\hat{\textbf{h}}^{(2)}; \textbf{W}^{(2)}_D)) + \hat{\textbf{h}}^{(2)})
\end{split}
\end{equation}
where $MSA_D$ is the self-attention module between the dimensions and $\phi(\cdot;\textbf{W}^{(2)}_D)$ is also a MLP layer with the shape of $64\times64$. 
The encoded individual representations $\textbf{e}_{xy}\in\mathbb{R}^{N\times64}$ is obtained by meanpooling $\textbf{h}^{(2)}\in\mathbb{R}^{N\times D\times64}$ in the second dimension. 

\begin{algorithm}[t]\small
\caption{Pseudo Code of the training of RLDE-AFL}\label{alg:pseudo}
\KwIn{Policy $\pi_\theta$, Critic $V_\psi$, Training instance Set $\mathcal{I}_\text{train}$}
\KwOut{Trained Policy $\pi_\theta$, Critic $V_\psi$}
\For{$epoch = 1$ \KwTo $Epoch$}{
    \For{$f \in \mathcal{I}_\text{train}$}{
        Initialize population $X_1$ and evaluation values $Y_1 = f(X_1)$\;
        \For{$t = 1$ \KwTo $T$}{
            Obtain state $s_t$ using $X_t$, $Y_t$ and $t$\;
            Determine actions $a_t = \pi_\theta(s_t)$\;
            Optimize $X_t$ using DE with configurations $a_t$ and obtain $X_{t+1}$, $Y_{t+1}$\;
            Calculate reward $r_t$ following Eq. (\ref{eq:reward})\;
            Collect the transition $<s_t, a_t, s_{t+1}, r_t>$\;
            \If{mod($t$, $n$) == 0}{
                \For{$k = 1$ \KwTo $\kappa$}{
                    Update $\pi_\theta$ and $V_\psi$ by \textbf{PPO} method\;
                }
            }
        }
    } 
} 
\end{algorithm}

\subsubsection{Time stamp feature.}
To take time stamp information into consideration when determining the actions, we embed $s_{time}$ into a 16-dimensional representation $e_{time} = \phi(s_{time};\textbf{W}_T)$ where $\phi(\cdot;\textbf{W}_T)$ is also a MLP layer with the shape of $1\times16$. Then repeat it for $N$ times to match the shape of $\textbf{e}_{xy}$: $\textbf{e}_{time} = \{e_{time}\}_{i=1}^N$. The decision vector $\textbf{dv}\in\mathbb{R}^{N\times(64+16)}$ is obtained as the concatenation of $\textbf{e}_{xy}$ and $\textbf{e}_{time}$: $\textbf{dv} = \text{Concat}(\textbf{e}_{xy}, \textbf{e}_{time})$. 

\begin{table*}[t]
\centering
\caption{The comparison results of the baselines on 10D testing problems.}
\label{tab:10Dres}
\resizebox{\textwidth}{!}{%
\begin{tabular}{c|cccccc|ccccccc}
 \hline
& \multicolumn{6}{c|}{Traditional DE Variants}      
& \multicolumn{7}{c}{RL-based DE Variants}
\\ \hline
& DE
& MadDE
& JDE21
& NL-SHADE-LBC
& AMCDE
& Random
& DE-DDQN
& DE-DQN
& LDE
& RL-HPSDE
& GLEET
& RL-DAS
& RLDE-AFL
\\ \hline
\multirow{8}{*}{}
F1
& \begin{tabular}[c]{@{}c@{}}4.584e1\\ $\pm$6.640e0\end{tabular} $+$
& \begin{tabular}[c]{@{}c@{}}2.614e1\\ $\pm$5.528e0\end{tabular} $+$
& \begin{tabular}[c]{@{}c@{}}3.051e1\\ $\pm$1.415e1\end{tabular} $+$
& \textbf{\begin{tabular}[c]{@{}c@{}}1.653e1\\ $\pm$4.289e0\end{tabular}} $\approx$
& \begin{tabular}[c]{@{}c@{}}3.161e1\\ $\pm$1.417e1\end{tabular} $+$
& \begin{tabular}[c]{@{}c@{}}3.032e1  \\ $\pm$7.238e0 \end{tabular} $+$
& \begin{tabular}[c]{@{}c@{}}4.004e1\\ $\pm$7.818e0\end{tabular} $+$
& \begin{tabular}[c]{@{}c@{}}3.118e2\\ $\pm$6.292e1\end{tabular} $+$
& \begin{tabular}[c]{@{}c@{}}3.471e1\\ $\pm$4.974e0\end{tabular} $+$
& \begin{tabular}[c]{@{}c@{}}7.457e1\\ $\pm$1.040e1\end{tabular} $+$
& \begin{tabular}[c]{@{}c@{}}2.815e1\\ $\pm$5.718e0\end{tabular} $+$
& \begin{tabular}[c]{@{}c@{}}3.400e1\\ $\pm$6.207e0\end{tabular} $+$
& \begin{tabular}[c]{@{}c@{}}1.811e1\\ $\pm$7.333e0\end{tabular}
\\
F2
& \begin{tabular}[c]{@{}c@{}}9.325e-1\\$\pm$3.203e-1\end{tabular} $+$
& \begin{tabular}[c]{@{}c@{}}1.170e-2\\ $\pm$6.815e-3\end{tabular} $+$
& \begin{tabular}[c]{@{}c@{}}4.893e-2\\ $\pm$2.101e-1\end{tabular} $+$
& \begin{tabular}[c]{@{}c@{}}1.846e-1\\ $\pm$1.371e-1\end{tabular} $+$
& \begin{tabular}[c]{@{}c@{}}4.423e0\\ $\pm$7.042e0\end{tabular} $+$
& \begin{tabular}[c]{@{}c@{}}1.521e-2  \\ $\pm$4.300e-2 \end{tabular} $+$
& \begin{tabular}[c]{@{}c@{}}4.669e-1\\ $\pm$3.250e0\end{tabular} $+$
& \begin{tabular}[c]{@{}c@{}}4.678e4\\ $\pm$2.687e4\end{tabular} $+$
& \begin{tabular}[c]{@{}c@{}}1.290e-1\\ $\pm$1.329e-1\end{tabular} $+$
& \begin{tabular}[c]{@{}c@{}}1.204e1\\ $\pm$4.692e0\end{tabular} $+$
& \begin{tabular}[c]{@{}c@{}}4.251e-1\\ $\pm$1.662e-1\end{tabular} $+$
& \begin{tabular}[c]{@{}c@{}}3.743e-2\\ $\pm$1.841e-2\end{tabular} $+$
& \textbf{\begin{tabular}[c]{@{}c@{}}2.525e-7
\\ $\pm$2.205e-7\end{tabular}}
\\
F3
& \begin{tabular}[c]{@{}c@{}}1.575e0\\ $\pm$3.954e-1\end{tabular} $+$
& \begin{tabular}[c]{@{}c@{}}4.395e-1\\ $\pm$2.559e-1\end{tabular} $\approx$
& \begin{tabular}[c]{@{}c@{}}7.377e-1\\ $\pm$4.520e-1\end{tabular} $+$
& \begin{tabular}[c]{@{}c@{}}2.993e-2\\ $\pm$6.775e-2\end{tabular} $-$
& \begin{tabular}[c]{@{}c@{}}2.218e0\\ $\pm$8.504e-1\end{tabular} $+$
& \begin{tabular}[c]{@{}c@{}}2.571e-1  \\ $\pm$1.616e-1 \end{tabular} $\approx$
& \textbf{\begin{tabular}[c]{@{}c@{}}5.655e-3\\ $\pm$3.999e-2\end{tabular}} $-$
& \begin{tabular}[c]{@{}c@{}}1.807e2\\ $\pm$6.459e1\end{tabular} $+$
& \begin{tabular}[c]{@{}c@{}}3.401e-1\\ $\pm$4.349e-1\end{tabular} $\approx$
& \begin{tabular}[c]{@{}c@{}}2.792e0\\ $\pm$1.591e0\end{tabular} $+$
& \begin{tabular}[c]{@{}c@{}}2.359e-2\\ $\pm$2.660e-2\end{tabular} $\approx$
& \begin{tabular}[c]{@{}c@{}}6.781e-1\\ $\pm$3.084e-1\end{tabular} $+$
& \begin{tabular}[c]{@{}c@{}}3.719e-1\\ $\pm$3.830e-1\end{tabular}
\\
F4
& \begin{tabular}[c]{@{}c@{}}8.196e0
\\ $\pm$4.098e0\end{tabular} $+$
& \begin{tabular}[c]{@{}c@{}}5.044e0\\ $\pm$4.435e-1\end{tabular} $+$
& \begin{tabular}[c]{@{}c@{}}3.359e0\\ $\pm$2.082e0\end{tabular} $+$
& \begin{tabular}[c]{@{}c@{}}4.557e0\\ $\pm$7.521e-1\end{tabular} $+$
& \begin{tabular}[c]{@{}c@{}}1.365e1\\ $\pm$2.285e1\end{tabular} $\approx$
& \begin{tabular}[c]{@{}c@{}}6.570e0  \\ $\pm$8.534e-1 \end{tabular} $+$
& \textbf{\begin{tabular}[c]{@{}c@{}}8.239e-1\\ $\pm$1.877e0\end{tabular} $-$}
& \begin{tabular}[c]{@{}c@{}}1.499e4\\ $\pm$7.657e3\end{tabular} $+$
& \begin{tabular}[c]{@{}c@{}}5.407e0\\ $\pm$1.995e0\end{tabular} $+$
& \begin{tabular}[c]{@{}c@{}}5.312e1\\ $\pm$3.281e1\end{tabular} $+$
& \begin{tabular}[c]{@{}c@{}}6.248e0\\ $\pm$9.377e-1\end{tabular} $+$
& \begin{tabular}[c]{@{}c@{}}5.417e0\\ $\pm$9.221e-1\end{tabular} $+$
& \begin{tabular}[c]{@{}c@{}}1.514e0\\ $\pm$1.093e0\end{tabular}
\\
F5
& \begin{tabular}[c]{@{}c@{}}7.506e0\\ $\pm$2.479e-1\end{tabular} $+$
& \begin{tabular}[c]{@{}c@{}}3.254e0\\ $\pm$6.338e-1\end{tabular} $+$
& \begin{tabular}[c]{@{}c@{}}6.780e0\\ $\pm$2.738e0\end{tabular} $+$
& \begin{tabular}[c]{@{}c@{}}6.944e0\\ $\pm$6.599e-1\end{tabular} $+$
& \begin{tabular}[c]{@{}c@{}}8.237e0\\ $\pm$7.712e0\end{tabular} $+$
& \begin{tabular}[c]{@{}c@{}}7.372e0  \\ $\pm$4.360e-1 \end{tabular} $+$
& \begin{tabular}[c]{@{}c@{}}2.431e0\\ $\pm$2.492e0\end{tabular} $\approx$
& \begin{tabular}[c]{@{}c@{}}1.258e4\\ $\pm$5.998e3\end{tabular} $+$
& \begin{tabular}[c]{@{}c@{}}7.088e0\\ $\pm$9.246e-1\end{tabular} $+$
& \begin{tabular}[c]{@{}c@{}}4.327e1\\ $\pm$2.368e1\end{tabular} $+$
& \begin{tabular}[c]{@{}c@{}}6.478e0\\ $\pm$7.257e-1\end{tabular} $+$
& \begin{tabular}[c]{@{}c@{}}4.363e0\\ $\pm$8.598e-1\end{tabular} $+$
& \textbf{\begin{tabular}[c]{@{}c@{}}2.407e0\\ $\pm$1.182e0\end{tabular}}
\\
F6
& \begin{tabular}[c]{@{}c@{}}1.380e4\\ $\pm$6.229e3\end{tabular} $+$
& \begin{tabular}[c]{@{}c@{}}8.980e2\\ $\pm$5.369e2\end{tabular} $+$
& \begin{tabular}[c]{@{}c@{}}6.242e2\\ $\pm$6.666e2\end{tabular} $+$
& \begin{tabular}[c]{@{}c@{}}9.828e1\\ $\pm$8.054e1\end{tabular} $-$
& \begin{tabular}[c]{@{}c@{}}2.839e3\\ $\pm$1.131e3\end{tabular} $+$
& \begin{tabular}[c]{@{}c@{}}2.588e3  \\ $\pm$1.023e3 \end{tabular} $+$
& \begin{tabular}[c]{@{}c@{}}4.560e1\\ $\pm$2.147e2\end{tabular} $-$
& \begin{tabular}[c]{@{}c@{}}5.294e5\\ $\pm$3.317e5\end{tabular} $+$
& \begin{tabular}[c]{@{}c@{}}2.206e2\\ $\pm$2.227e2\end{tabular} $\approx$
& \begin{tabular}[c]{@{}c@{}}2.621e3\\ $\pm$2.232e3\end{tabular} $+$
& \textbf{\begin{tabular}[c]{@{}c@{}}1.723e1\\ $\pm$1.643e1\end{tabular}} $-$
& \begin{tabular}[c]{@{}c@{}}1.387e3\\ $\pm$8.063e2\end{tabular} $+$
& \begin{tabular}[c]{@{}c@{}}1.530e2\\ $\pm$1.444e2\end{tabular}
\\

F7
& \begin{tabular}[c]{@{}c@{}}1.344e2\\ $\pm$2.775e1\end{tabular} $+$
& \begin{tabular}[c]{@{}c@{}}2.949e1\\ $\pm$1.125e1\end{tabular} $+$
& \begin{tabular}[c]{@{}c@{}}1.596e1\\ $\pm$8.448e0\end{tabular} $+$
& \begin{tabular}[c]{@{}c@{}}8.617e0\\ $\pm$4.953e0\end{tabular} $+$
& \begin{tabular}[c]{@{}c@{}}1.543e2\\ $\pm$3.598e1\end{tabular} $+$
& \begin{tabular}[c]{@{}c@{}}2.977e1  \\ $\pm$9.567e0 \end{tabular} $+$
& \begin{tabular}[c]{@{}c@{}}7.803e0\\ $\pm$6.936e0\end{tabular} $+$
& \begin{tabular}[c]{@{}c@{}}1.007e3\\ $\pm$1.446e3\end{tabular} $+$
& \begin{tabular}[c]{@{}c@{}}8.445e0\\ $\pm$5.847e0\end{tabular} $+$
& \begin{tabular}[c]{@{}c@{}}2.593e1\\ $\pm$2.523e1\end{tabular} $+$
& \textbf{\begin{tabular}[c]{@{}c@{}}1.525e-1\\ $\pm$1.147e-1\end{tabular}} $-$
& \begin{tabular}[c]{@{}c@{}}3.484e1\\ $\pm$9.974e0\end{tabular} $+$
& \begin{tabular}[c]{@{}c@{}}3.414e0\\ $\pm$2.532e0\end{tabular}
\\
F8
& \begin{tabular}[c]{@{}c@{}}1.369e3\\ $\pm$1.196e3\end{tabular} $+$
& \begin{tabular}[c]{@{}c@{}}8.770e1\\ $\pm$6.079e1\end{tabular} $+$
& \begin{tabular}[c]{@{}c@{}}8.144e0\\ $\pm$9.372e0\end{tabular} $+$
& \begin{tabular}[c]{@{}c@{}}2.285e0\\ $\pm$2.077e0\end{tabular} $+$
& \begin{tabular}[c]{@{}c@{}}7.474e3\\ $\pm$4.546e3\end{tabular} $+$
& \begin{tabular}[c]{@{}c@{}}5.914e6  \\ $\pm$1.772e6 \end{tabular} $+$
& \begin{tabular}[c]{@{}c@{}}1.211e4\\ $\pm$6.156e4\end{tabular} $+$ 
& \begin{tabular}[c]{@{}c@{}}2.924e7\\ $\pm$9.112e6\end{tabular} $+$
& \begin{tabular}[c]{@{}c@{}}3.627e0\\ $\pm$2.712e0\end{tabular} $+$
& \begin{tabular}[c]{@{}c@{}}1.992e5\\ $\pm$1.578e5\end{tabular} $+$
& \begin{tabular}[c]{@{}c@{}}2.942e1\\ $\pm$1.950e1\end{tabular} $+$
& \begin{tabular}[c]{@{}c@{}}7.466e1\\ $\pm$6.990e1\end{tabular} $+$
& \textbf{\begin{tabular}[c]{@{}c@{}}1.240e0\\ $\pm$1.575e0\end{tabular}}
\\ 
F9
& \begin{tabular}[c]{@{}c@{}}1.380e1\\ $\pm$3.017e0\end{tabular} $+$
& \begin{tabular}[c]{@{}c@{}}2.358e0\\ $\pm$8.985e-1\end{tabular} $+$
& \begin{tabular}[c]{@{}c@{}}5.347e0\\ $\pm$7.360e0\end{tabular} $\approx$
& \begin{tabular}[c]{@{}c@{}}1.023e-1\\ $\pm$1.827e-1\end{tabular} $-$
& \begin{tabular}[c]{@{}c@{}}1.561e1\\ $\pm$4.849e0\end{tabular} $+$
& \begin{tabular}[c]{@{}c@{}}4.783e0  \\ $\pm$1.832e0 \end{tabular} $+$
& \textbf{\begin{tabular}[c]{@{}c@{}}1.042e-6\\ $\pm$4.091e-7\end{tabular}} $-$
& \begin{tabular}[c]{@{}c@{}}9.133e2\\ $\pm$1.664e2\end{tabular} $+$
& \begin{tabular}[c]{@{}c@{}}5.508e-1\\ $\pm$9.579e-1\end{tabular} $-$
& \begin{tabular}[c]{@{}c@{}}7.408e1\\ $\pm$2.781e1\end{tabular} $+$
& \begin{tabular}[c]{@{}c@{}}1.254e0\\ $\pm$5.199e-1\end{tabular} $\approx$
& \begin{tabular}[c]{@{}c@{}}3.124e0\\ $\pm$1.405e0\end{tabular} $+$
& \begin{tabular}[c]{@{}c@{}}2.908e0\\ $\pm$3.724e0\end{tabular}
\\

F10
& \begin{tabular}[c]{@{}c@{}}3.599e-3\\ $\pm$8.465e-4\end{tabular} $+$
& \begin{tabular}[c]{@{}c@{}}8.888e-4\\ $\pm$4.605e-4\end{tabular} $+$
& \begin{tabular}[c]{@{}c@{}}4.771e-4\\ $\pm$3.167e-4\end{tabular} $+$
& \begin{tabular}[c]{@{}c@{}}7.050e-5\\ $\pm$5.500e-5\end{tabular} $\approx$
& \begin{tabular}[c]{@{}c@{}}3.025e-3\\ $\pm$1.414e-2\end{tabular} $-$
& \begin{tabular}[c]{@{}c@{}}4.252e-4  \\ $\pm$2.705e-4 \end{tabular} $+$
& \textbf{\begin{tabular}[c]{@{}c@{}}2.637e-8\\ $\pm$1.276e-7\end{tabular}} $-$
& \begin{tabular}[c]{@{}c@{}}1.193e1\\ $\pm$4.089e0\end{tabular} $+$
& \begin{tabular}[c]{@{}c@{}}2.611e-4\\ $\pm$1.704e-4\end{tabular} $+$
& \begin{tabular}[c]{@{}c@{}}2.080e-1\\ $\pm$1.373e-1\end{tabular} $+$
& \begin{tabular}[c]{@{}c@{}}1.996e-4\\ $\pm$1.118e-4\end{tabular} $+$
& \begin{tabular}[c]{@{}c@{}}1.570e-3\\ $\pm$4.325e-4\end{tabular} $+$
& \begin{tabular}[c]{@{}c@{}}6.642e-5\\ $\pm$3.046e-5\end{tabular}
\\

F11
& \begin{tabular}[c]{@{}c@{}}1.510e0\\ $\pm$3.539e-1\end{tabular} $+$
& \begin{tabular}[c]{@{}c@{}}1.435e0\\ $\pm$4.560e-1\end{tabular} $+$
& \begin{tabular}[c]{@{}c@{}}8.793e-1\\ $\pm$7.375e-1\end{tabular} $+$
& \begin{tabular}[c]{@{}c@{}}1.257e-1\\ $\pm$7.574e-2\end{tabular} $+$
& \begin{tabular}[c]{@{}c@{}}4.087e0\\ $\pm$1.314e0\end{tabular} $+$
& \begin{tabular}[c]{@{}c@{}}2.361e-2  \\ $\pm$3.254e-2 \end{tabular} $-$
& \textbf{\begin{tabular}[c]{@{}c@{}}3.786e-3\\ $\pm$1.742e-2\end{tabular}} $-$
& \begin{tabular}[c]{@{}c@{}}3.092e1\\ $\pm$6.330e0\end{tabular} $+$
& \begin{tabular}[c]{@{}c@{}}3.505e-1\\ $\pm$2.517e-1\end{tabular} $+$
& \begin{tabular}[c]{@{}c@{}}4.839e0\\ $\pm$1.536e0\end{tabular} $+$
& \begin{tabular}[c]{@{}c@{}}4.281e-1\\ $\pm$2.208e-1\end{tabular} $+$
& \begin{tabular}[c]{@{}c@{}}2.035e0\\ $\pm$4.582e-1\end{tabular} $+$
& \begin{tabular}[c]{@{}c@{}}7.239e-2\\ $\pm$9.005e-2\end{tabular}
\\

F12
& \begin{tabular}[c]{@{}c@{}}2.946e0\\ $\pm$4.763e-1\end{tabular} $+$
& \begin{tabular}[c]{@{}c@{}}1.138e0\\ $\pm$4.309e-1\end{tabular} $+$
& \begin{tabular}[c]{@{}c@{}}2.501e0\\ $\pm$4.347e-1\end{tabular} $+$
& \begin{tabular}[c]{@{}c@{}}2.200e0\\ $\pm$4.661e-1\end{tabular} $+$
& \begin{tabular}[c]{@{}c@{}}1.795e0\\ $\pm$1.308e0\end{tabular} $+$
& \begin{tabular}[c]{@{}c@{}}2.279e0  \\ $\pm$4.367e-1 \end{tabular} $+$
& \begin{tabular}[c]{@{}c@{}}2.279e0\\ $\pm$5.311e-1\end{tabular} $+$
& \begin{tabular}[c]{@{}c@{}}1.209e1\\ $\pm$2.158e0\end{tabular} $+$
& \begin{tabular}[c]{@{}c@{}}2.069e0\\ $\pm$3.717e-1\end{tabular} $+$
& \begin{tabular}[c]{@{}c@{}}4.256e0\\ $\pm$6.716e-1\end{tabular} $+$
& \begin{tabular}[c]{@{}c@{}}2.108e0\\ $\pm$4.858e-1\end{tabular} $+$
& \begin{tabular}[c]{@{}c@{}}9.524e-1\\ $\pm$2.673e-1\end{tabular} $+$
& \textbf{\begin{tabular}[c]{@{}c@{}}6.821e-1\\ $\pm$6.788e-1\end{tabular}} 
\\

F13
& \begin{tabular}[c]{@{}c@{}}1.610e0\\ $\pm$1.802e-1
\end{tabular} $+$
& \begin{tabular}[c]{@{}c@{}}8.774e-1\\ $\pm$1.687e-1\end{tabular} $+$
& \begin{tabular}[c]{@{}c@{}}4.500e-1\\ $\pm$2.202e-1\end{tabular} $-$
& \textbf{\begin{tabular}[c]{@{}c@{}}2.477e-1\\ $\pm$1.674e-1\end{tabular} $-$}
& \begin{tabular}[c]{@{}c@{}}1.514e0\\ $\pm$2.063e-1\end{tabular} $+$
& \begin{tabular}[c]{@{}c@{}}4.497e-1  \\ $\pm$2.208e-1 \end{tabular} $-$
& \begin{tabular}[c]{@{}c@{}}1.640e0\\ $\pm$3.319e-1\end{tabular} $+$
& \begin{tabular}[c]{@{}c@{}}7.034e3\\ $\pm$2.884e3\end{tabular} $+$
& \begin{tabular}[c]{@{}c@{}}1.491e0\\ $\pm$1.696e-1\end{tabular} $+$
& \begin{tabular}[c]{@{}c@{}}2.386e0\\ $\pm$3.024e-1\end{tabular} $+$
& \begin{tabular}[c]{@{}c@{}}1.091e0\\ $\pm$2.146e-1\end{tabular} $+$
& \begin{tabular}[c]{@{}c@{}}1.059e0\\ $\pm$2.536e-1\end{tabular} $+$
& \begin{tabular}[c]{@{}c@{}}6.569e-1\\ $\pm$2.343e-1\end{tabular}
\\

F14
& \begin{tabular}[c]{@{}c@{}}3.638e0\\ 1.000e0\end{tabular} $+$
& \textbf{\begin{tabular}[c]{@{}c@{}}5.860e-1\\ $\pm$2.263e-1\end{tabular}} $\approx$
& \begin{tabular}[c]{@{}c@{}}1.118e0\\ $\pm$7.399e-1\end{tabular} $\approx$
& \begin{tabular}[c]{@{}c@{}}6.666e-1\\ $\pm$1.251e-1\end{tabular} $\approx$
& \begin{tabular}[c]{@{}c@{}}1.705e0\\ $\pm$8.843e-1\end{tabular} $\approx$
& \begin{tabular}[c]{@{}c@{}}2.631e0  \\ $\pm$2.232e0 \end{tabular} $+$
& \begin{tabular}[c]{@{}c@{}}8.157e-1\\ $\pm$3.756e-1\end{tabular} $\approx$
& \begin{tabular}[c]{@{}c@{}}5.254e1\\ $\pm$1.298e1\end{tabular} $+$
& \begin{tabular}[c]{@{}c@{}}6.093e-1\\ $\pm$1.936e-1\end{tabular} $\approx$
& \begin{tabular}[c]{@{}c@{}}2.964e0\\ $\pm$1.844e0\end{tabular} $+$
& \begin{tabular}[c]{@{}c@{}}6.805e-1\\ $\pm$2.331e-1\end{tabular} $\approx$
& \begin{tabular}[c]{@{}c@{}}5.895e-1\\ $\pm$2.205e-1\end{tabular} $\approx$
& \begin{tabular}[c]{@{}c@{}}2.439e0\\ $\pm$2.302e0\end{tabular}
\\

F15
& \begin{tabular}[c]{@{}c@{}}1.837e0\\ $\pm$3.562e-1\end{tabular} $+$
& \begin{tabular}[c]{@{}c@{}}1.337e0\\ $\pm$3.141e-1\end{tabular} $+$
& \begin{tabular}[c]{@{}c@{}}1.414e0\\ $\pm$2.419e-1\end{tabular} $+$
& \begin{tabular}[c]{@{}c@{}}1.360e0\\ $\pm$2.802e-1\end{tabular} $+$
& \begin{tabular}[c]{@{}c@{}}1.930e0\\ $\pm$2.836e-1\end{tabular} $+$
& \begin{tabular}[c]{@{}c@{}}1.523e0  \\ $\pm$2.916e-1 \end{tabular} $+$
& \begin{tabular}[c]{@{}c@{}}1.313e0\\ $\pm$1.999e-1\end{tabular} $+$
& \begin{tabular}[c]{@{}c@{}}3.477e0\\ $\pm$9.457e-1\end{tabular} $+$
& \begin{tabular}[c]{@{}c@{}}1.374e0\\ $\pm$2.926e-1\end{tabular} $+$
& \begin{tabular}[c]{@{}c@{}}1.782e0\\ $\pm$3.684e-1\end{tabular} $+$
& \begin{tabular}[c]{@{}c@{}}1.252e0\\ $\pm$2.708e-1\end{tabular} $+$
& \begin{tabular}[c]{@{}c@{}}1.393e0\\ $\pm$2.121e-1\end{tabular} $+$
& \textbf{\begin{tabular}[c]{@{}c@{}}1.105e0\\ $\pm$4.564e-1\end{tabular}}
\\

F16
& \begin{tabular}[c]{@{}c@{}}3.992e1\\ $\pm$5.603e0\end{tabular} $+$
& \begin{tabular}[c]{@{}c@{}}4.178e1\\ $\pm$6.003e0\end{tabular} $+$
& \begin{tabular}[c]{@{}c@{}}3.841e1\\ $\pm$8.614e0\end{tabular} $+$
& \begin{tabular}[c]{@{}c@{}}3.581e1\\ $\pm$6.988e0\end{tabular} $+$
& \begin{tabular}[c]{@{}c@{}}5.055e1\\ $\pm$7.247e0\end{tabular} $+$
& \begin{tabular}[c]{@{}c@{}}3.706e1  \\ $\pm$7.623e0 \end{tabular} $+$
& \begin{tabular}[c]{@{}c@{}}4.177e1\\ $\pm$5.589e0\end{tabular} $+$
& \begin{tabular}[c]{@{}c@{}}1.686e2\\ $\pm$2.163e1\end{tabular} $+$
& \begin{tabular}[c]{@{}c@{}}4.138e1\\ $\pm$3.673e0\end{tabular} $+$
& \begin{tabular}[c]{@{}c@{}}6.822e1\\ $\pm$7.224e0\end{tabular} $+$
& \begin{tabular}[c]{@{}c@{}}4.157e1\\ $\pm$6.349e0\end{tabular} $+$
& \begin{tabular}[c]{@{}c@{}}4.411e1\\ $\pm$6.492e0\end{tabular} $+$
& \textbf{\begin{tabular}[c]{@{}c@{}}2.217e1\\ $\pm$6.183e0\end{tabular}}

\\ \hline
$+$ / $-$ / $\approx$
& 16 / 0 / 0
& 13 / 0 / 3
& 13 / 1 / 2
& 9 / 4 / 3
& 13 / 1 / 2
& 13 / 2 / 1
& 8 / 6 / 2
& 16 / 0 / 0
& 12 / 1 / 3
& 16 / 0 / 0
& 11 / 2 / 3
& 15 / 0 / 1
& N/A

\\ \hline
\end{tabular}%
}
\end{table*}

\subsubsection{Actor.}
Finally, the probabilities of selecting mutation operators $p_1$, probabilities of selecting crossover operators $p_2$, distribution of the parameters for mutation $\mathcal{N}(\mu_1, \sigma_1)$ and distribution of the parameters for crossover $\mathcal{N}(\mu_2, \sigma_2)$ for each individual are obtained through the MLP layers in the Actor separately as illustrated in the right of Figure~\ref{fig:network}. For instance, to select the mutation operator of the $i$-th individual, its decision vector is mapped to the 14-dimensional probability vector corresponding to the 14 candidate mutation operators: $p_{1,i} = \text{Softmax}(\phi(\textbf{dv}_i; \textbf{W}_M))$ by the MLP $\phi(\cdot;\textbf{W}_M)$ with shape $80\times32\times14$, and the index of the selected operator is sampled from Categorical$(p_{1,i})$. The parameters of the selected mutation operator is sampled from the normal distribution $a_i^{pc1}\sim\mathcal{N}(\phi(\textbf{dv}_i;\textbf{W}_{M\mu}), \text{Diag }\phi(\textbf{dv}_i;\textbf{W}_{M\sigma}))$ where $\phi(\cdot;\textbf{W}_{M\mu})$ and $\phi(\cdot;\textbf{W}_{M\sigma})$ are two MLP layers with the same shape of $80\times32 \times3$. 
It is worth noting that for parameter control, to uniformly control all operators which have diverse numbers of parameters, we pre-defined the maximum parameter numbers for mutation and crossover operators (in this paper, they are 3 and 2 respectively). The MLPs in Actor output 3 or 2 parameter for all operators. If the operator needs less parameters, the first few values would be used and the rest are ignored.

\subsubsection{Critic.}
For the critic $V_\psi$ parameterized by $\psi$, we calculate the value of an individual as $V_\psi(s_i) = \phi(\textbf{dv}_{i}; \textbf{W}_C)$ using a MLP with the shape of $80\times16\times8\times1$ and ReLU activation functions. The value of the population is the averaged value per individual $V_\psi(s) = \frac{1}{N}\sum_{i=1}^{N} V_\psi(s_{i})$.

\subsection{Training}

In this paper we use the $n$-step PPO~\cite{ppo} to train the policy in RLDE-AFL. As illustrated in Algorithm~\ref{alg:pseudo}, given a training problem instance set $\mathcal{I}_\text{train}$, for each epoch and each problem instance $f\in \mathcal{I}_\text{train}$, the DE algorithm first initializes a $N\times D$ population and evaluate it using $f$. For each generation $t$, the state $s_t$ is collected as mentioned in Section~\ref{sec:state}. The policy $\pi_\theta$ determines the action $a_t$ including the selected mutation and crossover operators and their corresponding parameters according to the state. With these configurations the DE algorithm optimize the population for one generation and obtain the next state $s_{t+1}$ and reward $r_t$. The transition $<s_t, a_t, s_{t+1}, r_t>$ is appended into a memory. For each $n$ generations, the actor $\pi_\theta$ and critic $V_\psi$ are updated for $\kappa$ steps using PPO manner.

\section{Experiment}\label{sec:Exp}
In this section, we discuss the following research questions: 
\textbf{RQ1:} How does the proposed RLDE-AFL perform on synthetic problem instances? 
\textbf{RQ2:} Can RLDE-AFL zero-shot to synthetic problems with expensive evaluation costs or different dimensions, as well as realistic problems? 
\textbf{RQ3:} How does the design of feature extractor affect RLDE-AFL's performance?
Below, we first introduce the experimental settings and then address RQ1$\sim$RQ3 respectively.

\subsection{Experimental Setup}

\subsubsection{Training setup.}
The following experiments are based on the MetaBox Benchmark~\cite{metabox} which provides the synthetic CoCo-BBOB benchmark~\cite{2009bbob}, the noisy-synthetic benchmark~\cite{2009bbob} and the Protein-Docking benchmark~\cite{protein}. In this paper we use 8 of the 24 problem instances with dimensions of 10 in the synthetic benchmark as training problem set $\mathcal{I}_\text{train}$ and the rest 16 problem instances as testing set (F1$\sim$F16). The detailed problem formulation and train-test split are provided in Appendix B. We train the policy for 100 epochs with a learning rate of 1e-3 and Adam optimizer. The PPO process is conducted $\kappa=3$ steps for every $n=10$ generation with discount factor $\gamma=0.99$.
For the DE optimization, the population size $N$ is set to 100, the maximum function evaluations is 20,000 thus the optimization horizon $T=200$. The searching space of all problem instances are $[-5, 5]^D$. All baselines in all experiments are run for 51 times. 
All experiments are run on Intel(R) Xeon(R) E5-2678 CPU and NVIDIA GeForce 1080Ti GPU with 32G RAM.

\begin{table*}[t]
\centering
\caption{The comparison results of the baselines on 20D testing problems. RL-DAS fails to generalize to 20D problems and is marked as ``/'' due to its problem dimension-dependent feature design.}
\label{tab:20Dres}
\resizebox{\textwidth}{!}{%
\begin{tabular}{c|cccccc|ccccccc}
 \hline
& \multicolumn{6}{c|}{Traditional DE Variants}      
& \multicolumn{7}{c}{RL-based DE Variants}
\\ \hline
& DE
& MadDE
& JDE21
& NL-SHADE-LBC
& AMCDE
& Random
& DE-DDQN
& DE-DQN
& LDE
& RL-HPSDE
& GLEET
& RL-DAS
& RLDE-AFL
\\ \hline
F1
& \begin{tabular}[c]{@{}c@{}}1.882e2\\ $\pm$1.563e1\end{tabular} $+$
& \begin{tabular}[c]{@{}c@{}}1.935e2\\ $\pm$1.528e1\end{tabular} $+$
& \begin{tabular}[c]{@{}c@{}}1.470e2\\ $\pm$4.598e1\end{tabular} $+$
& \begin{tabular}[c]{@{}c@{}}1.040e2\\ $\pm$1.598e1\end{tabular} $+$
& \begin{tabular}[c]{@{}c@{}}4.708e2\\ $\pm$1.105e2\end{tabular} $+$
& \begin{tabular}[c]{@{}c@{}}3.195e2\\ $\pm$7.503e1\end{tabular} $+$
& \begin{tabular}[c]{@{}c@{}}1.494e2\\ $\pm$2.528e1\end{tabular} $+$
& \begin{tabular}[c]{@{}c@{}}1.201e3\\ $\pm$2.781e2\end{tabular} $+$
& \begin{tabular}[c]{@{}c@{}}1.367e2\\ $\pm$1.689e1\end{tabular} $+$
& \begin{tabular}[c]{@{}c@{}}2.948e2\\ $\pm$4.318e1\end{tabular} $+$
& \begin{tabular}[c]{@{}c@{}}1.385e2\\ $\pm$1.444e1\end{tabular} $+$
& /
& \textbf{\begin{tabular}[c]{@{}c@{}}5.309e1\\ $\pm$1.559e1\end{tabular}}
\\

F2
& \begin{tabular}[c]{@{}c@{}}6.017e1\\ $\pm$9.446e0\end{tabular} $+$
& \begin{tabular}[c]{@{}c@{}}4.013e1\\ $\pm$4.998e0\end{tabular} $+$
& \begin{tabular}[c]{@{}c@{}}1.320e1\\ $\pm$1.426e1\end{tabular} $+$
& \begin{tabular}[c]{@{}c@{}}1.188e1\\ $\pm$4.803e0\end{tabular} $+$
& \begin{tabular}[c]{@{}c@{}}4.051e2\\ $\pm$1.661e2\end{tabular} $+$
& \begin{tabular}[c]{@{}c@{}}1.012e2\\ $\pm$6.252e1\end{tabular} $+$
& \begin{tabular}[c]{@{}c@{}}1.077e1\\ $\pm$1.382e1\end{tabular} $+$
& \begin{tabular}[c]{@{}c@{}}1.265e5\\ $\pm$5.919e4\end{tabular} $+$
& \begin{tabular}[c]{@{}c@{}}6.539e0\\ $\pm$3.717e0\end{tabular} $+$
& \begin{tabular}[c]{@{}c@{}}1.929e2\\ $\pm$3.054e1\end{tabular} $+$
& \begin{tabular}[c]{@{}c@{}}1.816e1\\ $\pm$4.907e0\end{tabular} $+$
& /
& \textbf{\begin{tabular}[c]{@{}c@{}}1.033e-1\\ $\pm$1.319e-1\end{tabular}}
\\

F3
& \begin{tabular}[c]{@{}c@{}}3.755e1\\ $\pm$7.877e0\end{tabular} $+$
& \begin{tabular}[c]{@{}c@{}}1.598e1\\ $\pm$3.362e0\end{tabular} $+$
& \begin{tabular}[c]{@{}c@{}}1.137e1\\ $\pm$7.579e0\end{tabular} $+$
& \textbf{\begin{tabular}[c]{@{}c@{}}2.798e0\\ $\pm$1.026e0\end{tabular} $-$}
& \begin{tabular}[c]{@{}c@{}}7.321e1\\ $\pm$1.616e1\end{tabular} $+$
& \begin{tabular}[c]{@{}c@{}}5.703e1\\ $\pm$2.681e1\end{tabular} $+$
& \begin{tabular}[c]{@{}c@{}}7.478e0\\ $\pm$7.017e0\end{tabular} $\approx$
& \begin{tabular}[c]{@{}c@{}}5.881e2\\ $\pm$1.345e2\end{tabular} $+$
& \begin{tabular}[c]{@{}c@{}}7.896e0\\ $\pm$2.971e0\end{tabular} $+$
& \begin{tabular}[c]{@{}c@{}}4.479e1\\ $\pm$2.070e1\end{tabular} $+$
& \begin{tabular}[c]{@{}c@{}}3.839e0\\ $\pm$1.451e0\end{tabular} $-$
& /
& \begin{tabular}[c]{@{}c@{}}6.225e0\\ $\pm$3.929e0\end{tabular} 
\\

F4
& \begin{tabular}[c]{@{}c@{}}6.885e1\\ $\pm$1.836e1\end{tabular} $+$
& \begin{tabular}[c]{@{}c@{}}1.039e2\\ $\pm$1.951e1\end{tabular} $+$
& \begin{tabular}[c]{@{}c@{}}4.256e1\\ $\pm$3.212e1\end{tabular} $+$
& \textbf{\begin{tabular}[c]{@{}c@{}}1.689e1\\ $\pm$6.453e-1\end{tabular} $+$}
& \begin{tabular}[c]{@{}c@{}}2.059e3\\ $\pm$2.578e3\end{tabular} $+$
& \begin{tabular}[c]{@{}c@{}}1.599e2\\ $\pm$8.383e1\end{tabular} $+$
& \begin{tabular}[c]{@{}c@{}}6.275e1\\ $\pm$4.567e1\end{tabular} $+$
& \begin{tabular}[c]{@{}c@{}}1.335e5\\ $\pm$3.857e4\end{tabular} $+$
& \begin{tabular}[c]{@{}c@{}}5.084e1\\ $\pm$3.004e1\end{tabular} $+$
& \begin{tabular}[c]{@{}c@{}}1.069e3\\ $\pm$6.218e2\end{tabular} $+$
& \begin{tabular}[c]{@{}c@{}}2.701e1\\ $\pm$1.516e1\end{tabular} $+$
& /
& \begin{tabular}[c]{@{}c@{}}1.745e1\\ $\pm$1.548e1\end{tabular}
\\

F5
& \begin{tabular}[c]{@{}c@{}}3.206e1\\ $\pm$5.580e0\end{tabular} $+$
& \begin{tabular}[c]{@{}c@{}}3.356e1\\ $\pm$1.030e1\end{tabular} $+$
& \begin{tabular}[c]{@{}c@{}}3.299e1\\ $\pm$2.766e1\end{tabular} $+$
& \begin{tabular}[c]{@{}c@{}}1.832e1\\ $\pm$7.480e-1\end{tabular} $+$
& \begin{tabular}[c]{@{}c@{}}8.553e3\\ $\pm$6.270e3\end{tabular} $+$
& \begin{tabular}[c]{@{}c@{}}1.161e2\\ $\pm$1.796e1\end{tabular} $+$
& \begin{tabular}[c]{@{}c@{}}1.966e1\\ $\pm$5.498e0\end{tabular} $+$
& \begin{tabular}[c]{@{}c@{}}7.805e4\\ $\pm$2.019e4\end{tabular} $+$
& \begin{tabular}[c]{@{}c@{}}1.854e1\\ $\pm$6.756e0\end{tabular} $+$
& \begin{tabular}[c]{@{}c@{}}6.568e2\\ $\pm$4.247e2\end{tabular} $+$
& \begin{tabular}[c]{@{}c@{}}1.875e1\\ $\pm$1.097e0\end{tabular} $+$
& /
& \textbf{\begin{tabular}[c]{@{}c@{}}1.423e1\\ $\pm$6.982e0\end{tabular} }
\\

F6
& \begin{tabular}[c]{@{}c@{}}2.281e5\\ $\pm$5.702e4\end{tabular} $+$
& \begin{tabular}[c]{@{}c@{}}2.856e4\\ $\pm$6.367e3\end{tabular} $+$
& \begin{tabular}[c]{@{}c@{}}1.070e4\\ $\pm$7.056e3\end{tabular} $+$
& \begin{tabular}[c]{@{}c@{}}4.260e3\\ $\pm$1.412e3\end{tabular} $\approx$
& \begin{tabular}[c]{@{}c@{}}2.006e5\\ $\pm$5.102e4\end{tabular} $+$
& \begin{tabular}[c]{@{}c@{}}3.678e4\\ $\pm$2.996e4\end{tabular} $+$
& \textbf{\begin{tabular}[c]{@{}c@{}}7.251e2\\ $\pm$3.789e3\end{tabular} $-$}
& \begin{tabular}[c]{@{}c@{}}1.937e6\\ $\pm$6.158e5\end{tabular} $+$
& \begin{tabular}[c]{@{}c@{}}4.689e3\\ $\pm$1.767e3\end{tabular} $\approx$
& \begin{tabular}[c]{@{}c@{}}1.050e5\\ $\pm$5.803e4\end{tabular} $+$
& \begin{tabular}[c]{@{}c@{}}2.992e3\\ $\pm$1.406e3\end{tabular} $-$
& /
& \begin{tabular}[c]{@{}c@{}}4.700e3\\ $\pm$1.913e3\end{tabular}
\\

F7
& \begin{tabular}[c]{@{}c@{}}2.863e2\\ $\pm$3.913e1\end{tabular} $+$
& \begin{tabular}[c]{@{}c@{}}7.694e1\\ $\pm$1.265e1\end{tabular} $+$
& \begin{tabular}[c]{@{}c@{}}6.426e1\\ $\pm$1.998e1\end{tabular} $+$
& \begin{tabular}[c]{@{}c@{}}3.377e1\\ $\pm$9.034e0\end{tabular} $-$
& \begin{tabular}[c]{@{}c@{}}3.440e2\\ $\pm$4.995e1\end{tabular} $+$
& \begin{tabular}[c]{@{}c@{}}9.865e1\\ $\pm$3.796e1\end{tabular} $+$
& \begin{tabular}[c]{@{}c@{}}3.162e1\\ $\pm$9.994e0\end{tabular} $-$
& \begin{tabular}[c]{@{}c@{}}1.682e3\\ $\pm$2.411e3\end{tabular} $+$
& \begin{tabular}[c]{@{}c@{}}4.300e1\\ $\pm$9.238e0\end{tabular} $-$
& \begin{tabular}[c]{@{}c@{}}1.268e2\\ $\pm$4.995e1\end{tabular} $+$
& \textbf{\begin{tabular}[c]{@{}c@{}}1.054e1\\ $\pm$5.289e0\end{tabular} $-$}
& /
& \begin{tabular}[c]{@{}c@{}}5.096e1\\ $\pm$1.108e1\end{tabular}
\\

F8
& \begin{tabular}[c]{@{}c@{}}7.019e4\\ $\pm$3.034e4\end{tabular} $+$
& \begin{tabular}[c]{@{}c@{}}1.437e5\\ $\pm$5.586e4\end{tabular} $+$
& \begin{tabular}[c]{@{}c@{}}1.186e3\\ $\pm$8.341e3\end{tabular} $\approx$
& \begin{tabular}[c]{@{}c@{}}1.550e2\\ $\pm$3.021e2\end{tabular} $+$
& \begin{tabular}[c]{@{}c@{}}9.262e6\\ $\pm$2.599e6\end{tabular} $+$
& \begin{tabular}[c]{@{}c@{}}1.302e6\\ $\pm$4.642e6\end{tabular} $+$
& \begin{tabular}[c]{@{}c@{}}5.609e6\\ $\pm$1.118e7\end{tabular} $+$
& \begin{tabular}[c]{@{}c@{}}1.682e8\\ $\pm$4.063e7\end{tabular} $+$
& \begin{tabular}[c]{@{}c@{}}7.870e0\\ $\pm$1.292e1\end{tabular} $+$
& \begin{tabular}[c]{@{}c@{}}5.778e6\\ $\pm$2.675e6\end{tabular} $+$
& \begin{tabular}[c]{@{}c@{}}6.007e3\\ $\pm$2.486e3\end{tabular} $+$
& /
& \textbf{\begin{tabular}[c]{@{}c@{}}4.590e0\\ $\pm$6.529e0\end{tabular} }
\\

F9
& \begin{tabular}[c]{@{}c@{}}1.014e2\\ $\pm$1.307e1\end{tabular} $+$
& \begin{tabular}[c]{@{}c@{}}1.156e2\\ $\pm$1.709e1\end{tabular} $+$
& \begin{tabular}[c]{@{}c@{}}3.106e1\\ $\pm$2.600e1\end{tabular} $\approx$
& \textbf{\begin{tabular}[c]{@{}c@{}}2.385e1\\ $\pm$5.755e0\end{tabular} $\approx$}
& \begin{tabular}[c]{@{}c@{}}5.438e2\\ $\pm$6.236e1\end{tabular} $+$
& \begin{tabular}[c]{@{}c@{}}2.517e2\\ $\pm$1.352e2\end{tabular} $+$
& \begin{tabular}[c]{@{}c@{}}1.292e2\\ $\pm$6.969e1\end{tabular} $+$
& \begin{tabular}[c]{@{}c@{}}1.861e3\\ $\pm$1.863e2\end{tabular} $+$
& \begin{tabular}[c]{@{}c@{}}6.266e1\\ $\pm$1.779e1\end{tabular} $+$
& \begin{tabular}[c]{@{}c@{}}3.772e2\\ $\pm$1.057e2\end{tabular} $+$
& \begin{tabular}[c]{@{}c@{}}2.697e1\\ $\pm$7.855e0\end{tabular} $\approx$
& /
& \begin{tabular}[c]{@{}c@{}}2.427e1\\ $\pm$2.134e1\end{tabular}
\\

F10
& \begin{tabular}[c]{@{}c@{}}1.422e-1\\ $\pm$3.063e-2\end{tabular} $+$
& \begin{tabular}[c]{@{}c@{}}1.182e-1\\ $\pm$3.807e-2\end{tabular} $+$
& \begin{tabular}[c]{@{}c@{}}6.561e-3\\ $\pm$5.119e-3\end{tabular} $+$
& \begin{tabular}[c]{@{}c@{}}2.250e-3\\ $\pm$9.769e-4\end{tabular} $+$
& \begin{tabular}[c]{@{}c@{}}4.823e0\\ $\pm$9.922e-1\end{tabular} $+$
& \begin{tabular}[c]{@{}c@{}}4.444e-1\\ $\pm$3.564e-1\end{tabular} $+$
& \begin{tabular}[c]{@{}c@{}}3.585e-1\\ $\pm$4.326e-1\end{tabular} $+$
& \begin{tabular}[c]{@{}c@{}}3.394e1\\ $\pm$7.914e0\end{tabular} $+$
& \begin{tabular}[c]{@{}c@{}}5.238e-3\\ $\pm$2.398e-3\end{tabular} $+$
& \begin{tabular}[c]{@{}c@{}}3.211e0\\ $\pm$1.500e0\end{tabular} $+$
& \begin{tabular}[c]{@{}c@{}}1.634e-2\\ $\pm$6.760e-3\end{tabular} $+$
& /
& \textbf{\begin{tabular}[c]{@{}c@{}}9.519e-4\\ $\pm$2.523e-4\end{tabular} }
\\

F11
& \begin{tabular}[c]{@{}c@{}}9.217e0\\ $\pm$1.420e0\end{tabular} $+$
& \begin{tabular}[c]{@{}c@{}}8.668e0\\ $\pm$1.432e0\end{tabular} $+$
& \begin{tabular}[c]{@{}c@{}}4.928e0\\ $\pm$2.658e0\end{tabular} $+$
& \begin{tabular}[c]{@{}c@{}}9.403e-1\\ $\pm$4.131e-1\end{tabular} $+$
& \begin{tabular}[c]{@{}c@{}}1.778e1\\ $\pm$2.345e0\end{tabular} $+$
& \begin{tabular}[c]{@{}c@{}}1.717e1\\ $\pm$4.165e0\end{tabular} $+$
& \begin{tabular}[c]{@{}c@{}}1.823e0\\ $\pm$1.160e0\end{tabular} $+$
& \begin{tabular}[c]{@{}c@{}}4.629e1\\ $\pm$4.713e0\end{tabular} $+$
& \begin{tabular}[c]{@{}c@{}}1.896e0\\ $\pm$7.897e-1\end{tabular} $+$
& \begin{tabular}[c]{@{}c@{}}1.558e1\\ $\pm$2.574e0\end{tabular} $+$
& \begin{tabular}[c]{@{}c@{}}3.284e0\\ $\pm$1.015e0\end{tabular} $+$
& /
& \textbf{\begin{tabular}[c]{@{}c@{}}4.828e-1\\ $\pm$4.071e-1\end{tabular}}
\\

F12
& \begin{tabular}[c]{@{}c@{}}5.729e0\\ $\pm$5.177e-1\end{tabular} $+$
& \begin{tabular}[c]{@{}c@{}}3.804e0\\ $\pm$4.603e-1\end{tabular} $\approx$
& \begin{tabular}[c]{@{}c@{}}4.888e0\\ $\pm$6.351e-1\end{tabular} $+$
& \begin{tabular}[c]{@{}c@{}}4.801e0\\ $\pm$4.913e-1\end{tabular} $+$
& \begin{tabular}[c]{@{}c@{}}7.365e0\\ $\pm$6.496e-1\end{tabular} $+$
& \textbf{\begin{tabular}[c]{@{}c@{}}2.504e-1\\ $\pm$9.903e-5\end{tabular} $-$}
& \begin{tabular}[c]{@{}c@{}}4.826e0\\ $\pm$3.498e-1\end{tabular} $+$
& \begin{tabular}[c]{@{}c@{}}2.052e1\\ $\pm$2.492e0\end{tabular} $+$
& \begin{tabular}[c]{@{}c@{}}4.339e0\\ $\pm$3.452e-1\end{tabular} $+$
& \begin{tabular}[c]{@{}c@{}}6.671e0\\ $\pm$7.936e-1\end{tabular} $+$
& \begin{tabular}[c]{@{}c@{}}4.461e0\\ $\pm$6.467e-1\end{tabular} $+$
& /
& \begin{tabular}[c]{@{}c@{}}3.301e0\\ $\pm$1.362e0\end{tabular}
\\

F13
& \begin{tabular}[c]{@{}c@{}}2.667e0\\ $\pm$1.361e-1\end{tabular} $+$
& \begin{tabular}[c]{@{}c@{}}2.348e0\\ $\pm$1.274e-1\end{tabular} $+$
& \textbf{\begin{tabular}[c]{@{}c@{}}9.153e-1\\ $\pm$1.933e-1\end{tabular} $-$}
& \begin{tabular}[c]{@{}c@{}}1.435e0\\ $\pm$1.930e-1\end{tabular} $+$
& \begin{tabular}[c]{@{}c@{}}1.836e1\\ $\pm$3.228e1\end{tabular} $+$
& \begin{tabular}[c]{@{}c@{}}6.684e0\\ $\pm$3.077e1\end{tabular} $+$
& \begin{tabular}[c]{@{}c@{}}2.670e0\\ $\pm$1.876e-1\end{tabular} $+$
& \begin{tabular}[c]{@{}c@{}}3.529e4\\ $\pm$1.060e4\end{tabular} $+$
& \begin{tabular}[c]{@{}c@{}}2.296e0\\ $\pm$1.037e-1\end{tabular} $+$
& \begin{tabular}[c]{@{}c@{}}1.224e1\\ $\pm$4.386e1\end{tabular} $+$
& \begin{tabular}[c]{@{}c@{}}2.098e0\\ $\pm$1.561e-1\end{tabular} $+$
& /
& \begin{tabular}[c]{@{}c@{}}1.109e0\\ $\pm$1.796e-1\end{tabular} 
\\

F14
& \begin{tabular}[c]{@{}c@{}}1.712e1\\ $\pm$8.110e0\end{tabular} $+$
& \begin{tabular}[c]{@{}c@{}}8.914e0\\ $\pm$3.003e-1\end{tabular} $+$
& \begin{tabular}[c]{@{}c@{}}9.162e0\\ $\pm$4.979e0\end{tabular} $+$
& \begin{tabular}[c]{@{}c@{}}8.113e0\\ $\pm$2.164e0\end{tabular} $+$
& \begin{tabular}[c]{@{}c@{}}4.562e1\\ $\pm$1.108e1\end{tabular} $+$
& \begin{tabular}[c]{@{}c@{}}1.304e1\\ $\pm$8.494e0\end{tabular} $+$
& \begin{tabular}[c]{@{}c@{}}1.229e1\\ $\pm$7.759e0\end{tabular} $+$
& \begin{tabular}[c]{@{}c@{}}7.855e1\\ $\pm$2.890e0\end{tabular} $+$
& \begin{tabular}[c]{@{}c@{}}7.721e0\\ $\pm$2.609e0\end{tabular} $+$
& \begin{tabular}[c]{@{}c@{}}1.490e1\\ $\pm$8.964e0\end{tabular} $+$
& \textbf{\begin{tabular}[c]{@{}c@{}}6.398e0\\ $\pm$3.650e0\end{tabular} $+$}
& /
& \begin{tabular}[c]{@{}c@{}}1.043e1\\ $\pm$6.426e0\end{tabular} 
\\
F15
& \begin{tabular}[c]{@{}c@{}}3.171e0\\ $\pm$5.202e-1\end{tabular} $+$
& \begin{tabular}[c]{@{}c@{}}2.478e0\\ $\pm$3.177e-1\end{tabular} $+$
& \begin{tabular}[c]{@{}c@{}}2.343e0\\ $\pm$3.829e-1\end{tabular} $\approx$
& \begin{tabular}[c]{@{}c@{}}2.415e0\\ $\pm$3.670e-1\end{tabular} $\approx$
& \begin{tabular}[c]{@{}c@{}}3.630e0\\ $\pm$5.854e-1\end{tabular} $+$
& \begin{tabular}[c]{@{}c@{}}2.522e0\\ $\pm$8.491e-1\end{tabular} $\approx$
& \begin{tabular}[c]{@{}c@{}}2.288e0\\ $\pm$3.562e-1\end{tabular} $\approx$
& \begin{tabular}[c]{@{}c@{}}4.156e0\\ $\pm$7.387e-1\end{tabular} $+$
& \begin{tabular}[c]{@{}c@{}}2.430e0\\ $\pm$3.896e-1\end{tabular} $+$
& \begin{tabular}[c]{@{}c@{}}2.709e0\\ $\pm$3.974e-1\end{tabular} $+$
& \textbf{\begin{tabular}[c]{@{}c@{}}2.112e0\\ $\pm$4.176e-1\end{tabular} $\approx$}
& /
& \begin{tabular}[c]{@{}c@{}}2.223e0\\ $\pm$4.864e-1\end{tabular}
\\
F16
& \begin{tabular}[c]{@{}c@{}}1.650e2\\ $\pm$1.069e1\end{tabular} $+$
& \begin{tabular}[c]{@{}c@{}}1.677e2\\ $\pm$8.599e0\end{tabular} $+$
& \begin{tabular}[c]{@{}c@{}}1.219e2\\ $\pm$1.844e1\end{tabular} $\approx$
& \begin{tabular}[c]{@{}c@{}}1.244e2\\ $\pm$1.091e1\end{tabular} $+$
& \begin{tabular}[c]{@{}c@{}}2.262e2\\ $\pm$2.286e1\end{tabular} $+$
& \begin{tabular}[c]{@{}c@{}}1.923e2\\ $\pm$3.426e1\end{tabular} $+$
& \begin{tabular}[c]{@{}c@{}}1.336e2\\ $\pm$1.358e1\end{tabular} $+$
& \begin{tabular}[c]{@{}c@{}}4.757e2\\ $\pm$4.756e1\end{tabular} $+$
& \begin{tabular}[c]{@{}c@{}}1.305e2\\ $\pm$8.811e0\end{tabular} $+$
& \begin{tabular}[c]{@{}c@{}}2.006e2\\ $\pm$1.658e1\end{tabular} $+$
& \begin{tabular}[c]{@{}c@{}}1.278e2\\ $\pm$1.276e1\end{tabular} $+$
& /
& \textbf{\begin{tabular}[c]{@{}c@{}}1.127e2\\ $\pm$1.966e1\end{tabular}}
\\ \hline
$+$ / $-$ / $\approx$
& 16 / 0 / 0
& 15 / 0 / 1
& 11 / 1 / 4
& 11 / 2 / 3
& 16 / 0 / 0
& 14 / 1 / 1
& 12 / 2 / 2
& 16 / 0 / 0
& 14 / 1 / 1
& 16 / 0 / 0
& 11 / 3 / 2
& /
& N/A
\\ \hline
\end{tabular}%
}
\end{table*}

\subsubsection{Baselines.} 
MetaBox integrates a large number of classic and advanced DE algorithms. In this paper we adopt vanilla DE~\cite{DE}, advanced DE variants MadDE~\cite{2021madde}, JDE21~\cite{2021jde}, NL-SHADE-LBC~\cite{2022nl-shade-lbc} and AMCDE~\cite{amcde} as traditional DE baselines. Besides, we include the random action RLDE-AFL without RL agent (denoted as ``Random'') to validate the effectiveness of RL training. For RL-based baselines, we adopt operator selection methods DE-DDQN~\cite{DEDDQN}, DE-DQN~\cite{DEDQN}, parameter control methods LDE~\cite{LDE}, GLEET~\cite{GLEET}, method that conducts both operator selection and parameter control RL-HPSDE~\cite{rlhpsde} and algorithm selection method RL-DAS~\cite{RLDAS}. The configurations of these baselines follow the setting in their original papers. RL-based baselines are trained on the same $\mathcal{I}_\text{train}$ for the same number of learning steps as RLDE-AFL for fair comparisons.

\subsection{Comparison on 10D Testing Set (RQ1)}

In this section we compare the optimization performance of our proposed RLDE-AFL with the baselines to answer RQ1. We train RLDE-AFL and RL-based baselines on the 8 problem instance $\mathcal{I}_\text{train}$ and test them on F1$\sim$F16. The mean and standard deviation of the 51 runs for all baselines are presented in Table~\ref{tab:10Dres}. The `$+$', `$-$' and `$\approx$' indicate the statistical test results that the performance of RLDE-AFL is better, worse and no significant difference with competing methods respectively according to the Wilcoxon rank-sum test at the 0.05 significance level.
The results show that:
\begin{enumerate}
    \item Our RLDE-AFL outperforms all traditional and RL-based baselines, achieving the state-of-the-art performance, which validates the effectiveness of the proposed method.  
    \item RLDE-AFL significantly surpasses the random action baseline which reveals the advantage of RL-based configuration policy and validates the effectiveness of the RL training. Besides, the random action baselines outperforms DE which reveals that integrating diverse operators would enhance the robustness of algorithm and lead to certain performance even using random operator selection and parameters.
    \item Compared to traditional baselines, the RL based configuration policy in RLDE-AFL shows superior performance than human-crafted adaptive mechanisms, indicating that RL agents could not only relieve the expertise dependence in adaptive mechanism designs, but also present promising optimization performance. 
    \item Compared to RL-based baselines using human-crafted features, the superior performance of RLDE-AFL validates the effectiveness of the NeurELA based feature extraction. Besides, instead of exclusively selecting operators or controlling parameters, including both operator selection and parameter control for individuals in action space could minimize the expertise dependence, unleash the behavior diversity, and hence obtain better performance.
\end{enumerate}

\subsection{Zero-shot Generalization Performance (RQ2)}

In this section, we conduct the zero-shot generalization to 20D F1$\sim$F16 synthetic problems, expensive problems and realistic problems to answer RQ2. Specifically, we use the RLDE-AFL model trained with 10D synthetic training set $\mathcal{I}_\text{train}$ to the aforementioned testing problem sets without further tuning. The RL-based baselines are also zero-shot generalized in the same way as RLDE-AFL.

\subsubsection{Zero-shot to 20D problems.}

This section generalizes baselines from 10D training problem instances to 20D instances. As presented in Table~\ref{tab:20Dres}, RLDE-AFL maintains the advantages and surpasses the baselines, showing the promising generalization across problem dimensions. 
The cross-dimension attention in NeurELA based feature extractor enables the agent to process populations with different dimensions uniformly and successfully capture the optimization information, while some baselines such as RL-DAS fails in generalization due to its dimension-dependent state design.

\subsubsection{Zero-shot to expensive problems.}

\begin{figure}[t]
\centering
\includegraphics[width=\columnwidth]{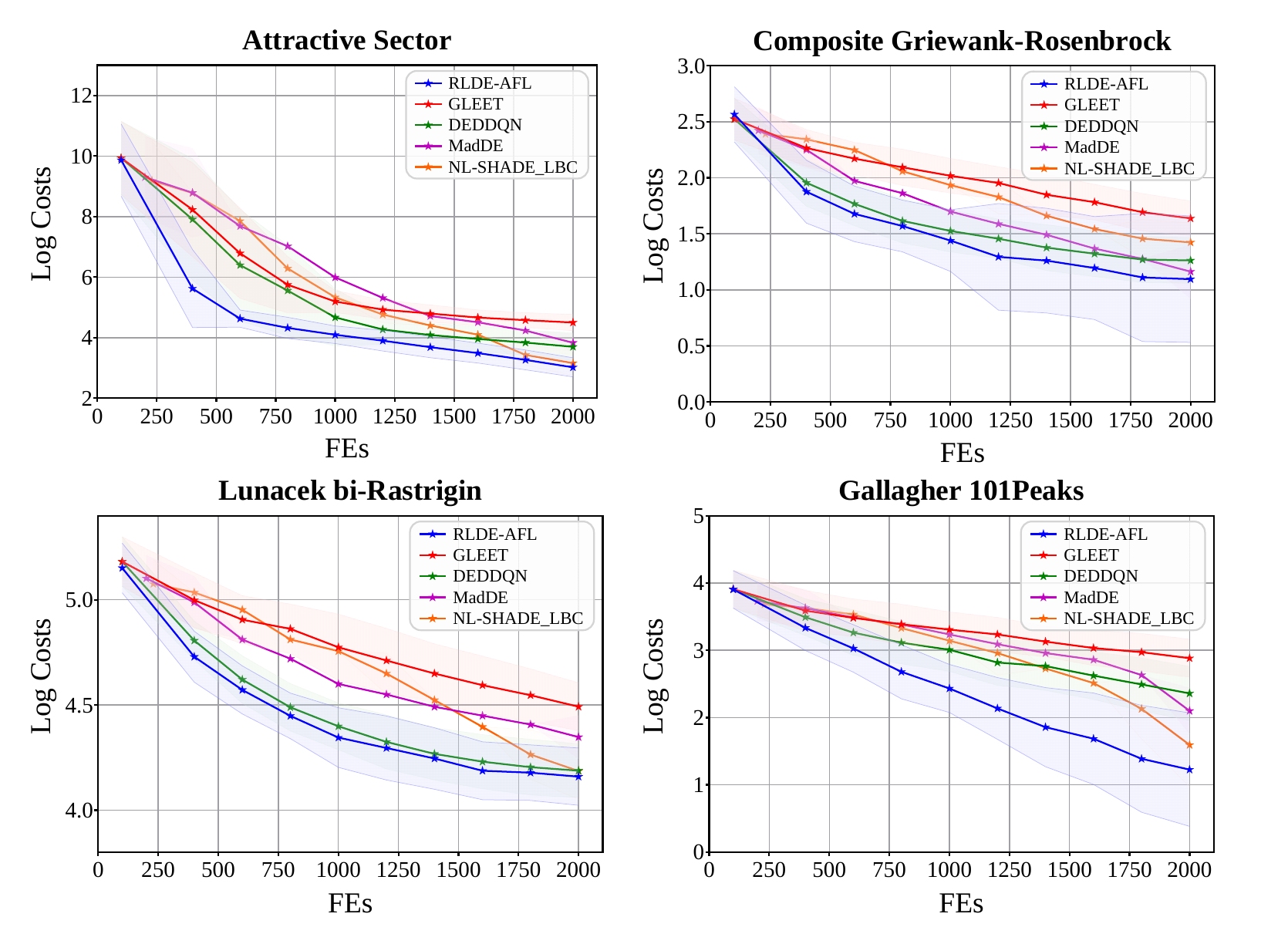}
\caption{The optimization curves of RLDE-AFL and baselines on the four problems with 2,000 function evaluations.}
\label{fig:2kres}
\end{figure}

In this section we reduce the maximum function evaluations from 20,000 to 2,000 and hence shorten the optimization horizon to simulate the expensive optimization scenarios in which maximum function evaluations are more limited. The problem instances in these scenarios would be much more challenging. We compare RLDE-AFL and 4 best baselines in 10D comparison: MadDE, NL-SHADE-LBC, DE-DDQN and GLEET. 
Figure~\ref{fig:2kres} presents the optimization curves of the baselines under 4 10D synthetic problem instances in MetaBox: Attractive Sector (F2), Composite Griewank-Rosenbrock (F12), Lunacek bi-Rastrigin (F16) and Gallagher 101Peaks (training problem). The results show that RLDE-AFL has faster convergence speed in expensive scenarios and hence surpasses the 4 baselines.
The results infers the potential of applying the RLDE-AFL models pre-trained on non-expensive problem instances to solve expensive problems, saving training cost while obtaining promising performance.

\subsubsection{Zero-shot to realistic problems.}

The experiments above are conducted on synthetic problems which are in the same distribution as synthetic training problem instances. To further validate the generalization ability of RLDE-AFL on out-of-distribution problems, we employ the Protein-Docking problems~\cite{protein} provided in MetaBox benchmark which containing 280 12D protein-docking task instances. Due to the computationally expensive evaluations, the maximum number of function evaluations is set to 500. The default search range for the optimization is $[-5, 5]$. We zero-shot RLDE-AFL and RL-based baselines DE-DDQN and GLEET trained on 10D synthetic problem instances to the protein-docking benchmark. 
The best objective value AEI~\cite{metabox,ma2024toward} results between RLDE-AFL and the 4 best baselines MadDE, NL-SHADE-LBC, DE-DDQN and GLEET are presented in Figure~\ref{fig:protein}, the details of the AEI metric is provided in Appendix C. 
On protein-docking problem instances, RLDE-AFL shows competitive performance with GLEET and surpasses DE-DDQN, MadDE and NL-SHADE-LBC, revealing the remarkable out-of-distribution generalization ability of RLDE-AFL. 

\begin{figure}[t]
\centering
\includegraphics[width=0.9\columnwidth]{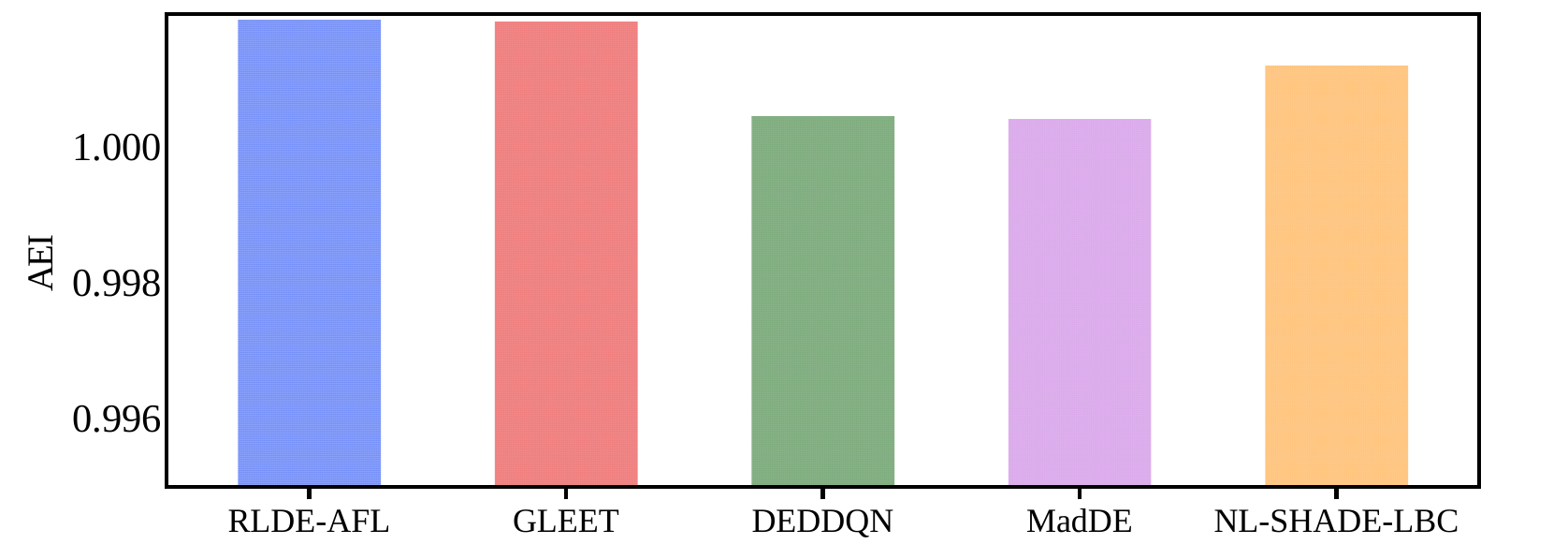}
\caption{The AEI scores of RLDE-AFL and the baselines on the protein-docking realistic problem set.}
\label{fig:protein}
\end{figure}

\begin{table}[t]
\centering
\caption{The averaged accumulated rewards of the ablated baselines and RLDE-AFL.}
\resizebox{0.98\columnwidth}{!}{%
\begin{tabular}{c|ccccc}
\hline
 &
  w/o Time &
  w/o ME &
  MLP &
  HandCraft &
  RLDE-AFL
  \\ \hline
 \begin{tabular}[c]{@{}c@{}}10D\\problems\end{tabular} &
  \begin{tabular}[c]{@{}c@{}}9.497e-1\\ $\pm$1.231e-2\end{tabular} &
  \begin{tabular}[c]{@{}c@{}}9.442e-1\\ $\pm$1.135e-2\end{tabular} &
  \begin{tabular}[c]{@{}c@{}}9.451e-1\\ $\pm$1.419e-2\end{tabular} &
  \begin{tabular}[c]{@{}c@{}}9.512e-1\\ $\pm$9.478e-3\end{tabular} &
  \textbf{\begin{tabular}[c]{@{}c@{}}9.645e-1\\ $\pm$6.577e-3\end{tabular}}
\\ \hline
 \begin{tabular}[c]{@{}c@{}}20D\\problems\end{tabular} &
  \begin{tabular}[c]{@{}c@{}}9.289e-01\\ $\pm$1.500e-02\end{tabular} &
  \begin{tabular}[c]{@{}c@{}}9.207e-01\\ $\pm$1.592e-02\end{tabular} &
  \begin{tabular}[c]{@{}c@{}}9.274e-01\\ $\pm$1.414e-02\end{tabular} &
  \begin{tabular}[c]{@{}c@{}}9.317e-01\\ $\pm$1.243e-02\end{tabular} &
  \textbf{\begin{tabular}[c]{@{}c@{}}9.334e-01\\ $\pm$1.295e-02\end{tabular}}
\\ \hline
 \begin{tabular}[c]{@{}c@{}}Expensive\\problems\end{tabular} &
  \begin{tabular}[c]{@{}c@{}}8.836e-01\\ $\pm$3.069e-02\end{tabular} &
  \begin{tabular}[c]{@{}c@{}}8.713e-01\\ $\pm$2.950e-02\end{tabular} &
  \begin{tabular}[c]{@{}c@{}}8.847e-01\\ $\pm$3.079e-02\end{tabular} &
  \begin{tabular}[c]{@{}c@{}}8.714e-01\\ $\pm$2.767e-02\end{tabular} &
  \textbf{\begin{tabular}[c]{@{}c@{}}8.856e-01\\ $\pm$2.494e-02\end{tabular}}
\\ \hline
 \begin{tabular}[c]{@{}c@{}}Realistic\\problems\end{tabular} &
  \begin{tabular}[c]{@{}c@{}}6.279e-01\\ $\pm$1.475e-03\end{tabular} &
  \begin{tabular}[c]{@{}c@{}}5.664e-01\\ $\pm$1.300e-03\end{tabular} &
  \begin{tabular}[c]{@{}c@{}}6.603e-01\\ $\pm$1.535e-03\end{tabular} &
  \begin{tabular}[c]{@{}c@{}}6.565e-01\\ $\pm$1.519e-03\end{tabular} &
  \textbf{\begin{tabular}[c]{@{}c@{}}7.056e-01\\ $\pm$1.608e-03\end{tabular}}
\\ \hline
\end{tabular}%
}
\label{tab:abla}
\end{table}

\subsection{Ablation Study (RQ3)}

In order to verify the effectiveness of the neural network based feature extractor, we conduct the ablation studies on the state design and feature extractor. We first remove the time stamp feature in the state and use NeurELA features only (denoted as ``w/o Time''). Then we further ablate the mantissa-exponent representation and apply the min-max normalization adopted in original NeurELA (denoted as ``w/o ME'') for evaluation value normalization. Next we replace the attention modules in NeurELA by simple MLPs to validate the necessity of the information sharing between solutions and dimensions (denoted as ``MLP''). Lastly we remove the whole NeurELA and instead use human-crafted optimization features proposed in GLEET as the states (denoted as ``HandCraft''). The boxplots of the accumulated rewards of RLDE-AFL and the ablated baselines on 10D, 20D, expensive and realistic testing problems are presented in Table~\ref{tab:abla}. 
The RLDE-AFL variant without time stamp feature slightly underperforms RLDE-AFL.The RL agent informed with optimization progress information could accordingly adjust the configuration along the optimization and hence achieve better performance. 
The baseline ``w/o ME'' uses min-max normalization within the populations which fails in distinguishing different problem instances and different optimization progresses, leading to poor performance. It confirms the necessity of the mantissa-exponent representation which not only normalizes the scales but also provides the agent exact performance of the individuals thereby supporting purposeful configuration. The lower performance of the MLP baseline validates the significance of the information sharing between individuals and dimensions in comprehensive decision-making. The superior performance of RLDE-AFL over the baseline with hand-crafted features emphasizes the effectiveness of the neural network based feature extractor, which relieves the human effort burden while acquiring better performance.


\section{Conclusion}\label{sec:conclu}

This paper proposed a novel MetaBBO approach, RLDE-AFL, with RL-based policy, generalizable NeurELA based feature extractor and diverse mutation and crossover operators. By integrating the attention-based learnable feature extraction module with mantissa-exponent based fitness representation to encode the population and evaluation values into expressive optimization states, we relieved the expertise dependency in feature design and enhanced the generalization ability. We formulated the optimization process as a MDP and incorporated a comprehensive algorithm configuration space including the integrated diverse DE operators into a RL-aided algorithm configuration paradigm. Experimental results verified that the proposed RLDE-AFL not only showed promising optimization performance on in-distribution synthetic problems, but also presented robust generalization ability across problem dimensions and optimization horizons, as well as the out-of-distribution realistic BBO scenarios. In-depth analysis on state feature extraction further validated the effectiveness of the NeurELA module and the necessity of co-training the proposed feature learning module.

\begin{acks}
    This work was supported in part by the National Natural Science Foundation of China No. 62276100, in part by the Guangdong Provincial Natural Science Foundation for Outstanding Youth Team Project No. 2024B1515040010, in part by the Guangdong Natural Science Funds for Distinguished Young Scholars No. 2022B1515020049, and in part by the TCL Young Scholars Program.
\end{acks}

\bibliographystyle{ACM-Reference-Format}
\bibliography{sample_sigconf}

\newpage
\onecolumn
\appendix
\section{Operator Collection}

In this section we list the names, formulations, descriptions and configuration spaces of the integrated DE mutation and crossover operators in Table~\ref{tab:operator}.

\begin{table*}[h]
\centering
\caption{The names, formulations, descriptions and configuration spaces of the integrated DE mutation and crossover operators.}
\label{tab:operator}
\resizebox{\textwidth}{!}{%
\begin{tabular}{c|c|c|l|l}
\hline
Index & Operator & Formula & Description & Parameter\\ \hline
1
& rand/1
& $u = x_{r1} + F\cdot(x_{r2} - x_{r3})$
& \begin{tabular}[l]{@{}l@{}}Generate exploratory trail solutions using 3 randomly selected individuals $x_{r\cdot}$.\end{tabular} 
& $F\in[0,1]$
\\ \hline
2
& best/1
& $u = x^* + F\cdot(x_{r1} - x_{r2})$ 
& \begin{tabular}[l]{@{}l@{}}Generate exploitative trail solutions by leveraging the best individual $x^*$ from the current population and random individuals.\end{tabular} 
& $F\in[0,1]$
\\ \hline
3
& rand/2
& $u = x_{r1} + F\cdot(x_{r2} - x_{r3}) + F\cdot(x_{r4} - x_{r5})$
& Generate exploratory trail solution with 5 randomly selected individuals, enhancing the exploration of rand/1.
& $F\in[0,1]$
\\ \hline
4
& best/2
& $u = x^* + F\cdot(x_{r1} - x_{r2}) + F\cdot(x_{r3} - x_{r4})$
& Generate exploitative trail solutions by leveraging the best individual and 4 random individuals, which is more exploratory than best/1.
& $F\in[0,1]$
\\ \hline
5
& current-to-rand/1
& $u = x + F\cdot(x_{r1} - x) + F\cdot(x_{r2} - x_{r3})$
& \begin{tabular}[l]{@{}l@{}}Generate trail solution by leveraging the current individual $x$ and random individuals.\end{tabular} 
& $F\in[0,1]$
\\ \hline
6
& current-to-best/1
& $u = x + F\cdot(x^* - x) + F\cdot(x_{r1} - x_{r2})$
& Generate trail solution by leveraging the current individual $x$ and the best solution $x^*$.
& $F\in[0,1]$
\\ \hline
7
& rand-to-best/1
& $u = x_{r1} + F\cdot(x^* - x_{r2}) + F\cdot(x_{r3} - x_{r4})$
& Generate trail solution by leveraging the random inviduals $x_{r\cdot}$ and the best solution $x^*$ from the current population.
& $F\in[0,1]$
\\ \hline
8
& current-to-pbest/1~\cite{MDE_pBX}
& $u = x + F\cdot(x^*_p - x) + F\cdot(x_{r1} - x_{r2})$
& \begin{tabular}[l]{@{}l@{}}A variant of current-to-best/1, replace the best individual with the individual randomly selected from the top $p$\% individuals in the \\population $x^*_p$ for better exploration.\end{tabular} 
& \begin{tabular}[l]{@{}l@{}}$F\in[0,1]$\\$p\in[0,1]$\end{tabular} 
\\ \hline
9
& current-to-pbest/1 + archive~\cite{JADE} 
& $u = x + F\cdot(x^*_p - x) + F\cdot(x_{r1} - \Tilde{x}_{r2})$
& \begin{tabular}[l]{@{}l@{}}A variant of current-to-pbest/1, replace the last random individual with a individual \\randomly selected from the union of the population and archive $\Tilde{x}_{r\cdot}$ to further enhance the exploration.\end{tabular}
& \begin{tabular}[l]{@{}l@{}}$F\in[0,1]$\\$p\in[0,1]$\end{tabular} 
\\ \hline
10
& current-to-rand/1 + archive~\cite{2021madde}
& $u = x + F\cdot(x_{r1} - \Tilde{x}_{r2})$
& \begin{tabular}[l]{@{}l@{}}A variant of current-to-rand/1, which also replaces the last random individual with a individual randomly selected from the union of \\the population and archive $\Tilde{x}_{r\cdot}$.\end{tabular}
& $F\in[0,1]$
\\ \hline
11
& weighted-rand-to-pbest/1~\cite{2021madde}
& $u = F \cdot x_{r1} + F\cdot F_{a} \cdot(x^*_p - x_{r2})$
& \begin{tabular}[l]{@{}l@{}}A variant of rand-to-best/1, replace the best individual with the individual randomly selected from the top $p$\% individuals in the \\population $x^*_p$ and add a new parameter to enhance the behaviour diversity.\end{tabular}
& \begin{tabular}[l]{@{}l@{}}$F\in[0,1]$\\$F_a\in[0,1]$\\$p\in[0,1]$\end{tabular} 
\\ \hline
12 
& ProDE-rand/1~\cite{PRO-DE}
& $u = x_{p1} + F\cdot(x_{p2} - x_{p3})$
& \begin{tabular}[l]{@{}l@{}}A variant of rand/1, which takes the distances between individuals into consideration and assigns close individuals larger selection \\probabilities, enhancing the exploitation of rand/1.\end{tabular}
& $F\in[0,1]$
\\ \hline
13
& HARDDE-current-to-pbest/2~\cite{harde}
& $u = x + F\cdot(x^*_p - x) + F_1\cdot(x_{r1} - \Tilde{x}_{r2}) + F_1\cdot(x_{r1} - \Tilde{x}_{r3})$
& \begin{tabular}[l]{@{}l@{}}A variant of current-to-pbest/1, replace the last random individual with a individual randomly selected from the union of the \\population and archive $\Tilde{x}_{r2}$, and add a solution difference using the individual $\Tilde{x}_{r3}$ randomly selected from the union of the population \\and former individual archive.\end{tabular}
& \begin{tabular}[l]{@{}l@{}}$F\in[0,1]$\\$F_{1}\in[0,1]$\\$p\in[0,1]$\end{tabular}
\\ \hline
14
& TopoDE-rand/1~\cite{TopoMut-DE}
& $u = x_{nb} + F\cdot(x_{r2} - x_{r3})$
& A variant of rand/1, use the best individual $x_{nb}$ in the nearest individuals to replace the first random individual in rand/1.
& \begin{tabular}[l]{@{}l@{}}$F\in[0,1]$\end{tabular}
\\ \hline
 \hline
1
& binomial crossover
& $v_j = \begin{cases}
u_j & \text{If }rand_j < Cr \text{ or }j = jrand;\\
x_j & \text{Otherwise.}
\end{cases}\quad j=1,2,\dots,D$
& \begin{tabular}[l]{@{}l@{}}Basic crossover operator that randomly selects values from trail solution or parent solution and uses a random index $jrand$ to ensure \\that the generated individual is different from the parent.\end{tabular}
& $Cr\in[0,1]$
\\ \hline
2
& exponential crossover
& $v_j = \begin{cases}
u_j & \text{If }j\in \left\{ \left< n \right> _D,\left< n+1 \right> _D,...,\left< n+L-1 \right> _D \right\}; \\
x_j & \text{Otherwise.}
\end{cases}\quad j=1,2,\dots,D$
& \begin{tabular}[l]{@{}l@{}}Basic crossover operator that randomly selects a random length segment of the parent and cover the parent values in this segment \\using trail solution values.\end{tabular}
& $Cr\in[0,1]$
\\ \hline
3
& p-binomial crossover~\cite{2021madde}
& $v_j = \begin{cases}
u_j & \text{If }rand_j < Cr \text{ or }j = jrand;\\
\Tilde{x}^p_j & \text{Otherwise.}
\end{cases}\quad j=1,2,\dots,D$
& \begin{tabular}[l]{@{}l@{}}A variant of binomial crossover which replaces the parent individual with the individual randomly selected from the top $p$\% individuals \\in the population $x^*_p$.\end{tabular} 
& \begin{tabular}[l]{@{}l@{}}$Cr\in[0,1]$\\$p\in[0,1]$\end{tabular} 
\\ \hline
\end{tabular}%
}
\end{table*}

\section{Training \& Testing Problem Set}

\begin{figure*}[h]
    \centering
    \subfloat[$f_{1}$]{
    \label{fig:func_1}
    \includegraphics[height=0.15\textwidth]{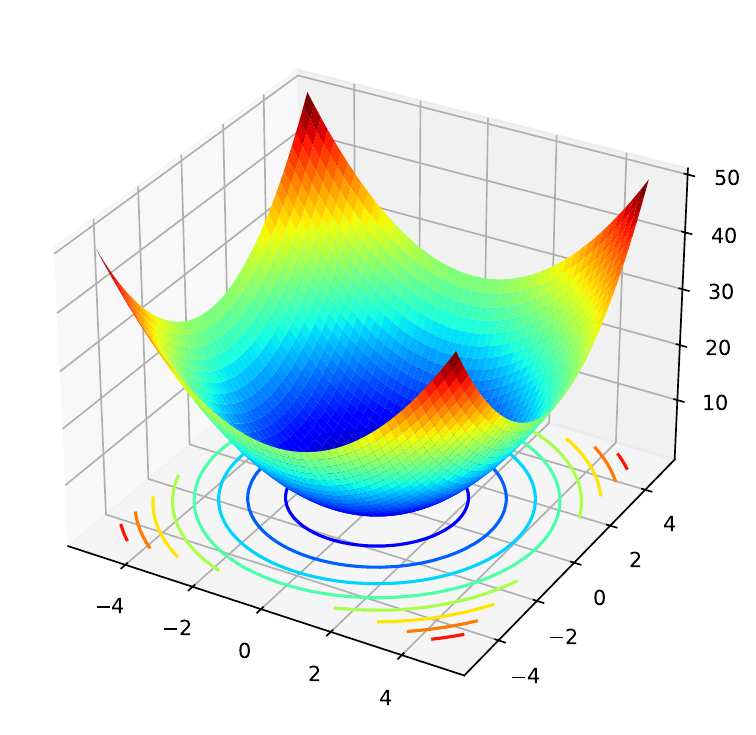}
    }
    \hfill
    \subfloat[$f_{2}$]{
    \label{fig:func_2}
    \includegraphics[height=0.15\textwidth]{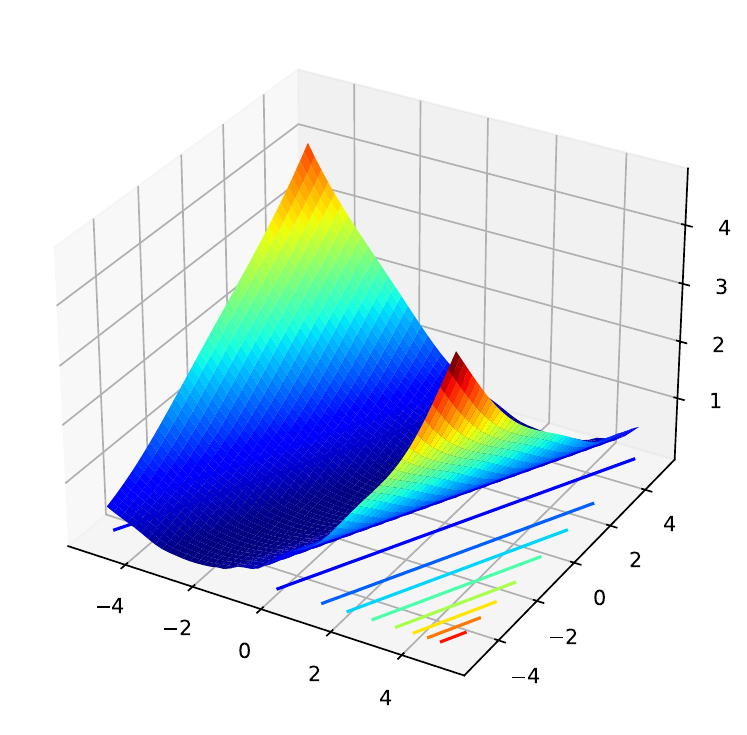}
    }
    \hfill
    \subfloat[$f_{3}$]{
    \label{fig:func_3}
    \includegraphics[height=0.15\textwidth]{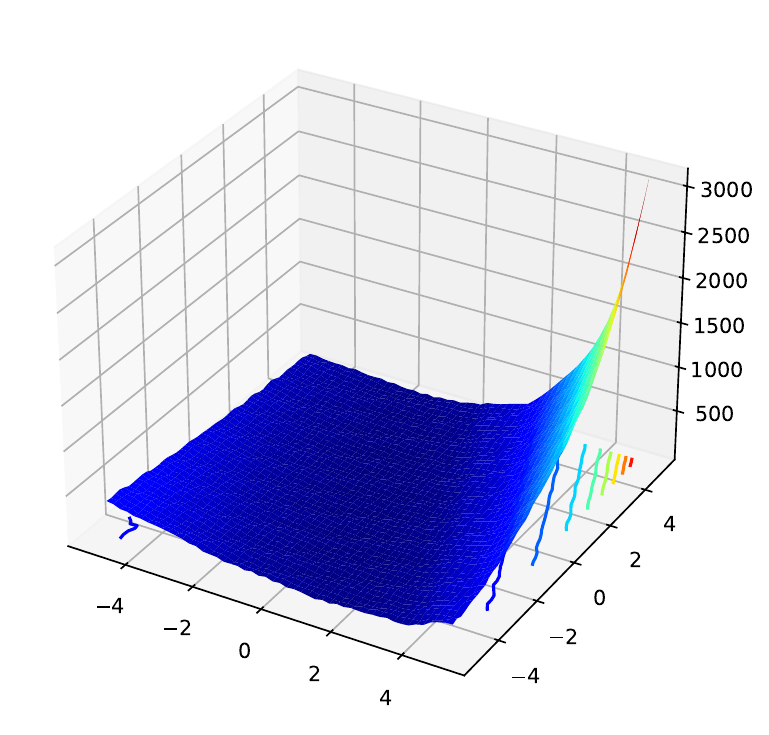}
    }
    \hfill
    \subfloat[$f_{5}$]{
    \label{fig:func_5}
    \includegraphics[height=0.15\textwidth]{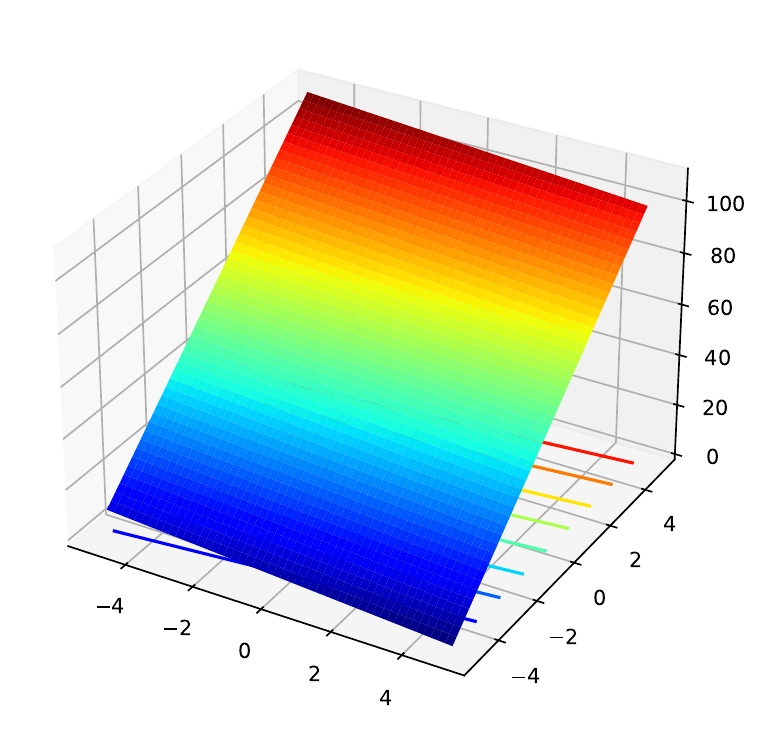}
    }
    \\
    \subfloat[$f_{15}$]{
    \label{fig:func_15}
    \includegraphics[height=0.15\textwidth]{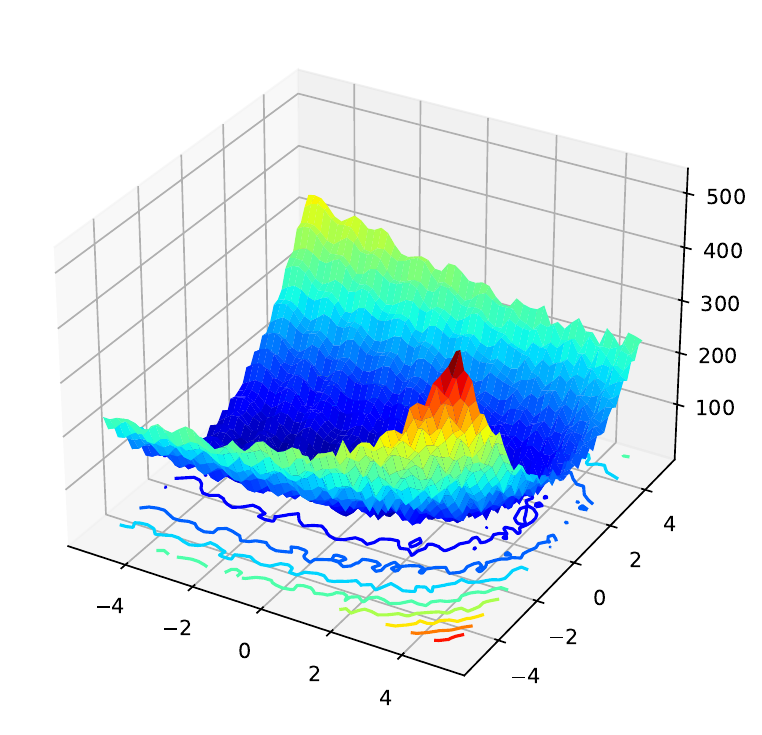}
    }
    \hfill
    \subfloat[$f_{16}$]{
    \label{fig:func_16}
    \includegraphics[height=0.15\textwidth]{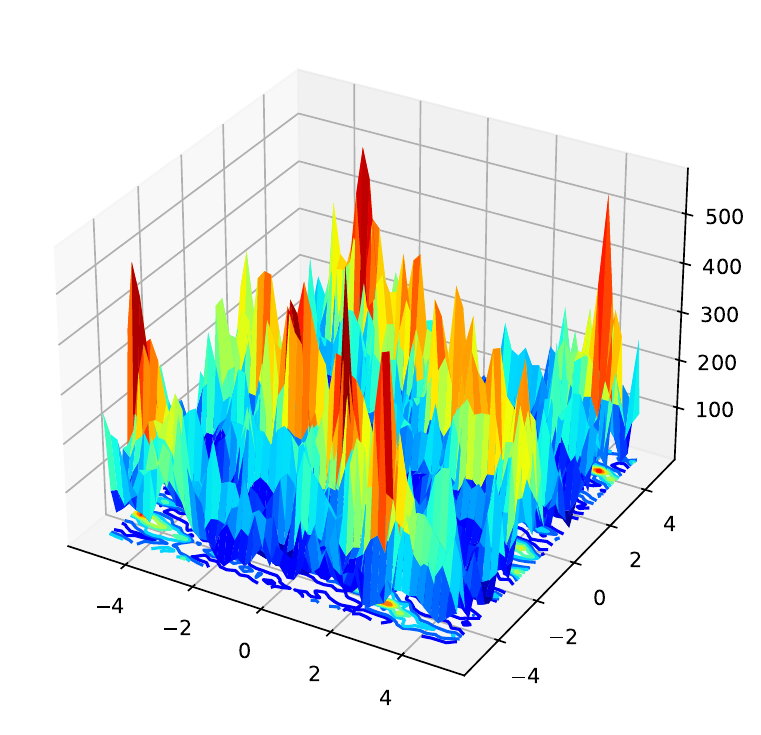}
    }
    \hfill
    \subfloat[$f_{17}$]{
    \label{fig:func_17}
    \includegraphics[height=0.15\textwidth]{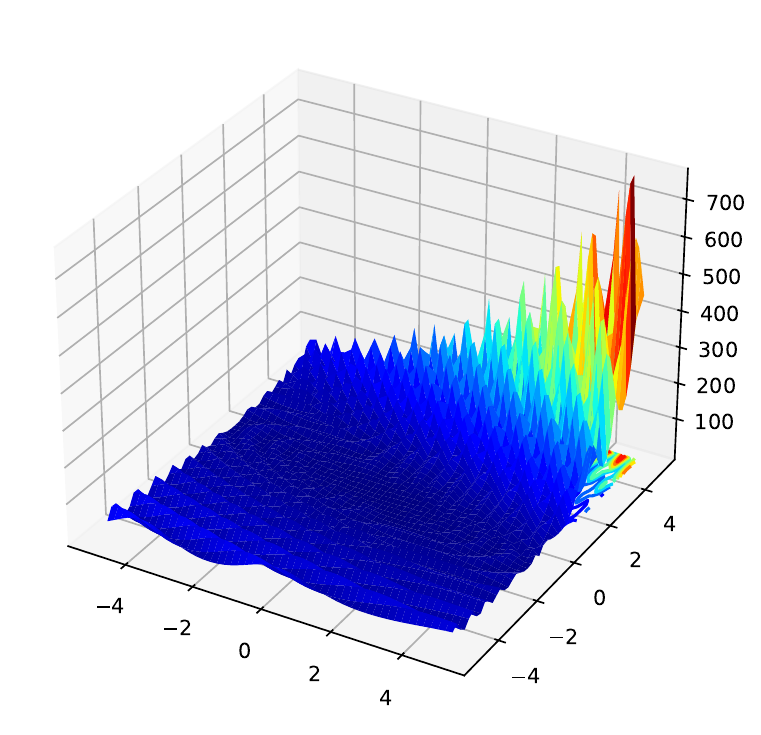}
    }
    \hfill
    \subfloat[$f_{21}$]{
    \label{fig:func_21}
    \includegraphics[height=0.15\textwidth]{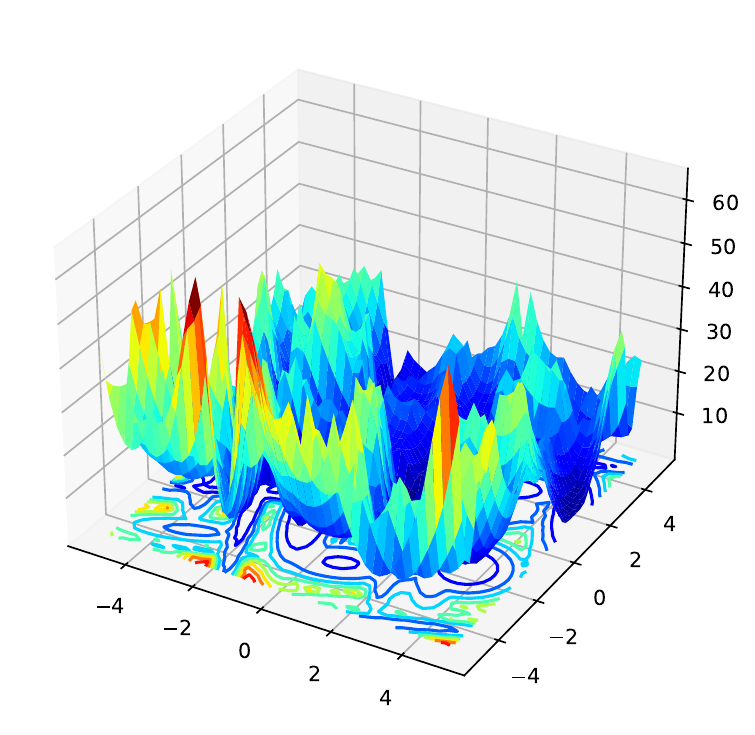}
    }
    \caption{Fitness landscapes of functions in the training set when dimension is set to 2. }
    \label{fig:train_func}
\end{figure*}

\begin{figure*}[h]
    \centering
    \subfloat[F1: $f_{4}$, Buche-Rastrigin]{
    \label{fig:func_4}
    \includegraphics[height=0.15\textwidth]{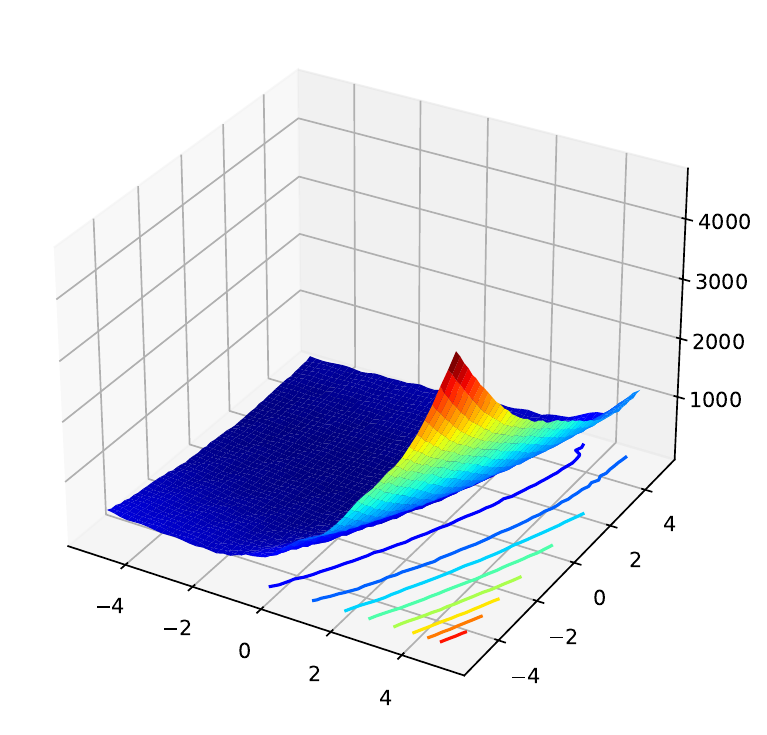}
    }
    \hfill
    \subfloat[F2: $f_{6}$, Attractive Sector]{
    \label{fig:func_6}
    \includegraphics[height=0.15\textwidth]{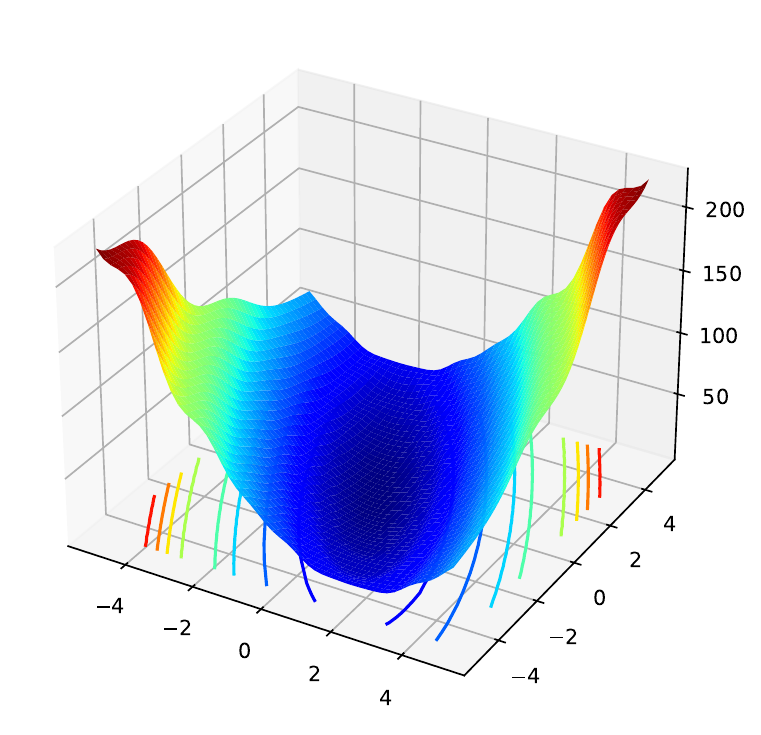}
    }
    \hfill
    \subfloat[F3: $f_{7}$, Step Ellipsoidal]{
    \label{fig:func_7}
    \includegraphics[height=0.15\textwidth]{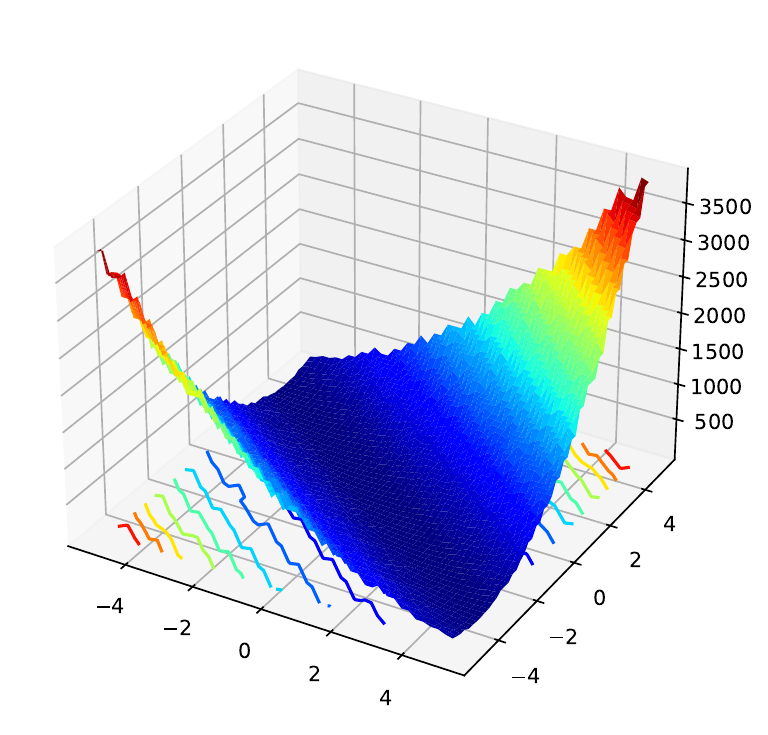}
    }
    \hfill
    \subfloat[F4: $f_{8}$, original Rosenbrock Function]{
    \label{fig:func_8}
    \includegraphics[height=0.15\textwidth]{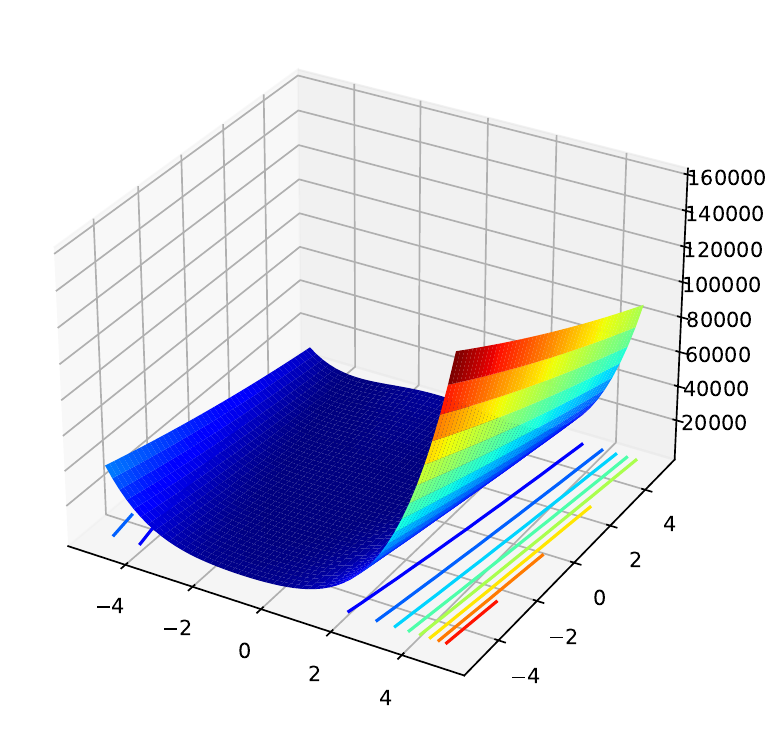}
    }
    \\
    \subfloat[F5: $f_{9}$, rotated Rosenbrock Function]{
    \label{fig:func_9}
    \includegraphics[height=0.15\textwidth]{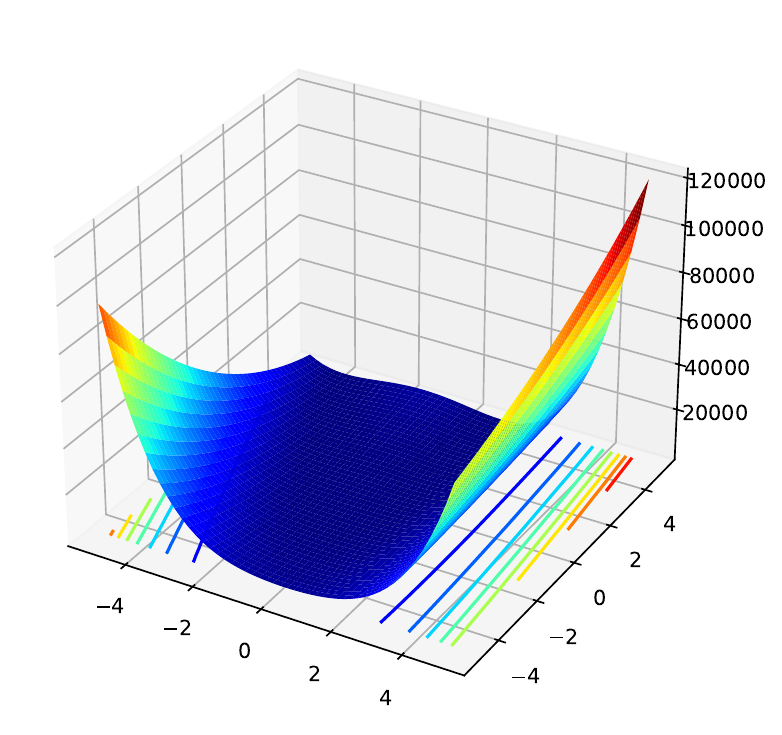}
    }
    \hfill
    \subfloat[F6: $f_{10}$, Ellipsoidal]{
    \label{fig:func_10}
    \includegraphics[height=0.15\textwidth]{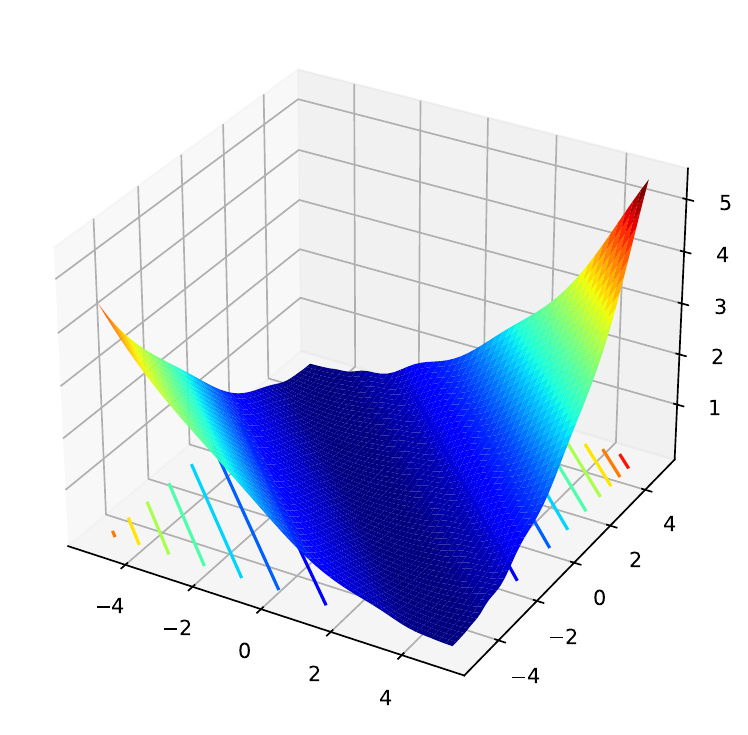}
    }
    \hfill
    \subfloat[F7: $f_{11}$, Discus]{
    \label{fig:func_11}
    \includegraphics[height=0.15\textwidth]{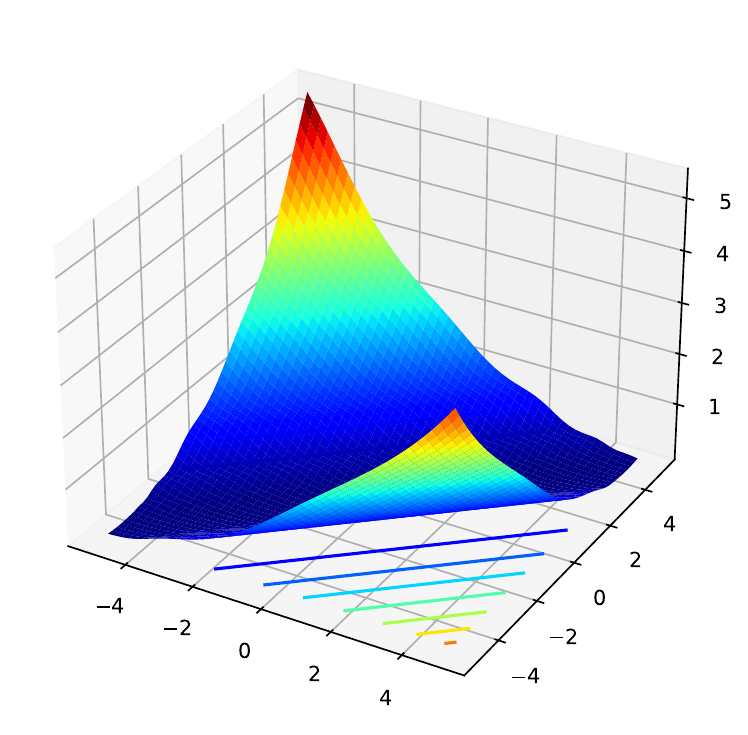}
    }
    \hfill
    \subfloat[F8: $f_{12}$, Bent Cigar]{
    \label{fig:func_12}
    \includegraphics[height=0.15\textwidth]{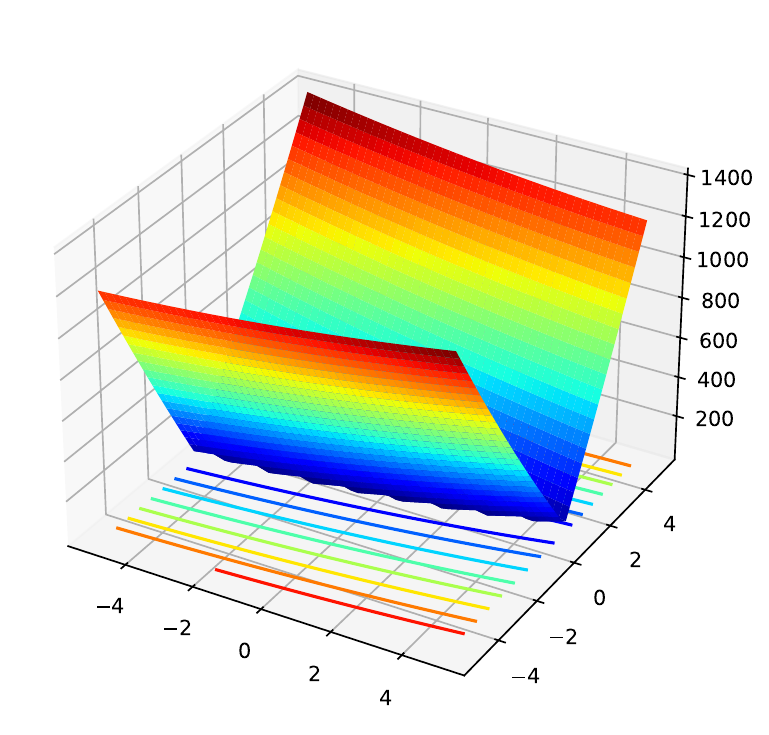}
    }
    \\
    \subfloat[F9: $f_{13}$, Sharp Ridge]{
    \label{fig:func_13}
    \includegraphics[height=0.15\textwidth]{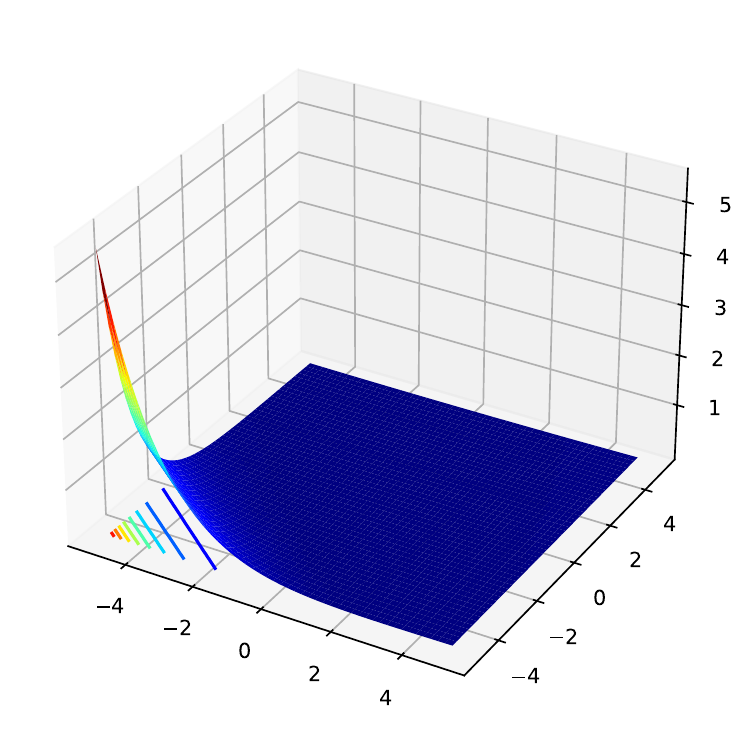}
    }
    \hfill
    \subfloat[F10: $f_{14}$, Different Powers]{
    \label{fig:func_14}
    \includegraphics[height=0.15\textwidth]{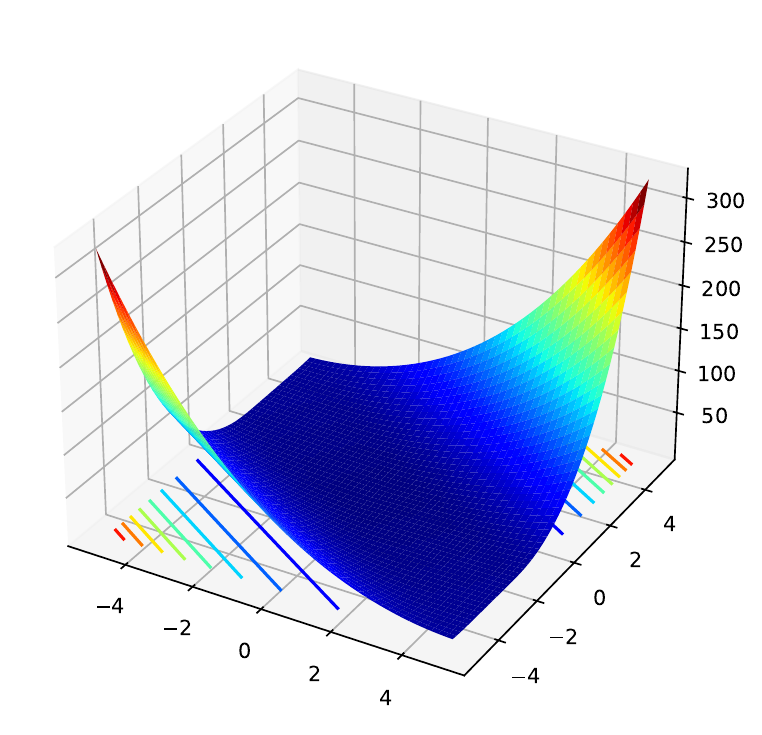}
    }
    \hfill
    \subfloat[F11: $f_{18}$, moderately ill-conditioned Schaffers F7]{
    \label{fig:func_18}
    \includegraphics[height=0.15\textwidth]{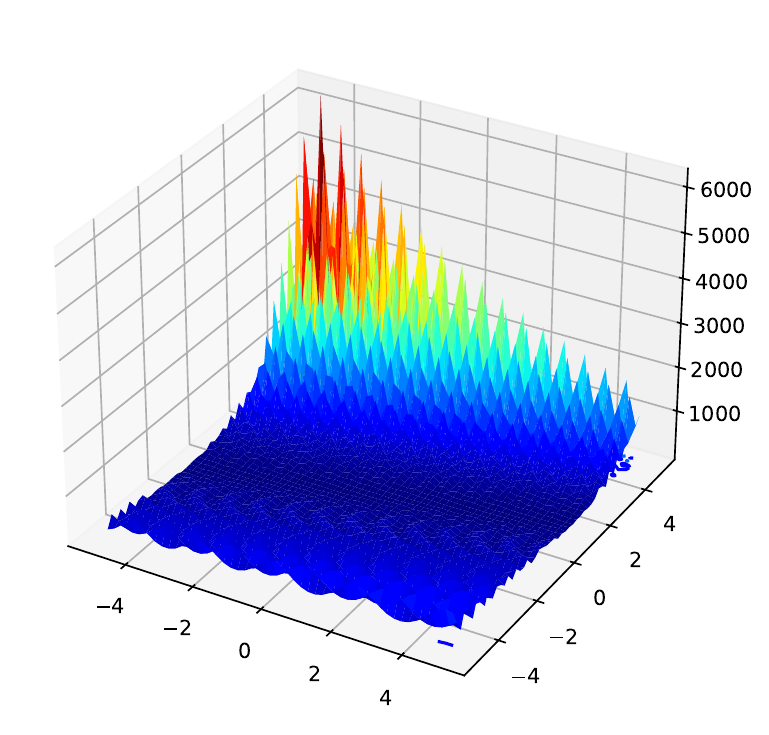}
    }
    \hfill
    \subfloat[F12: $f_{19}$, Composite Griewank-Rosenbrock]{
    \label{fig:func_19}
    \includegraphics[height=0.15\textwidth]{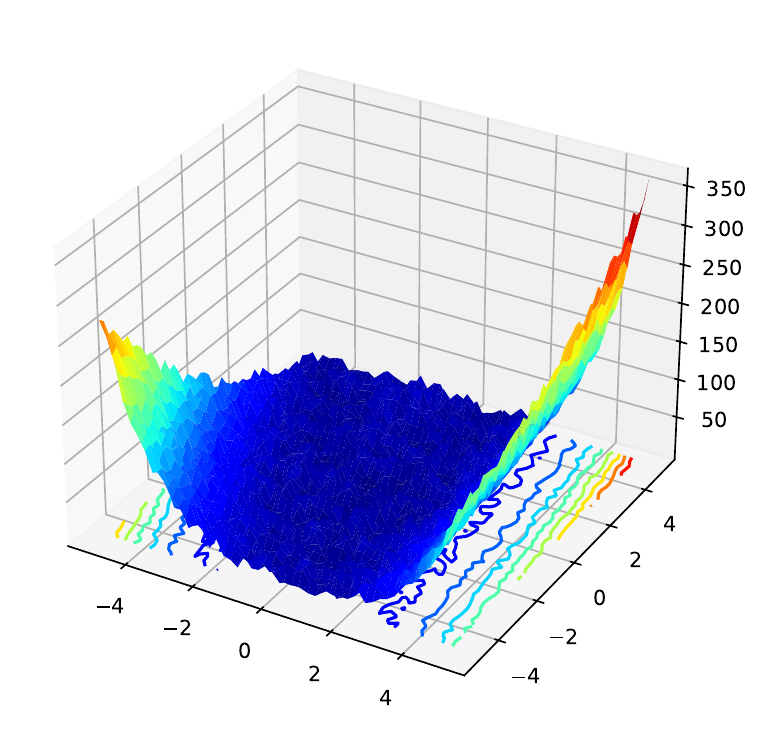}
    }
    \\
    \subfloat[F13: $f_{20}$, Schwefel]{
    \label{fig:func_20}
    \includegraphics[height=0.15\textwidth]{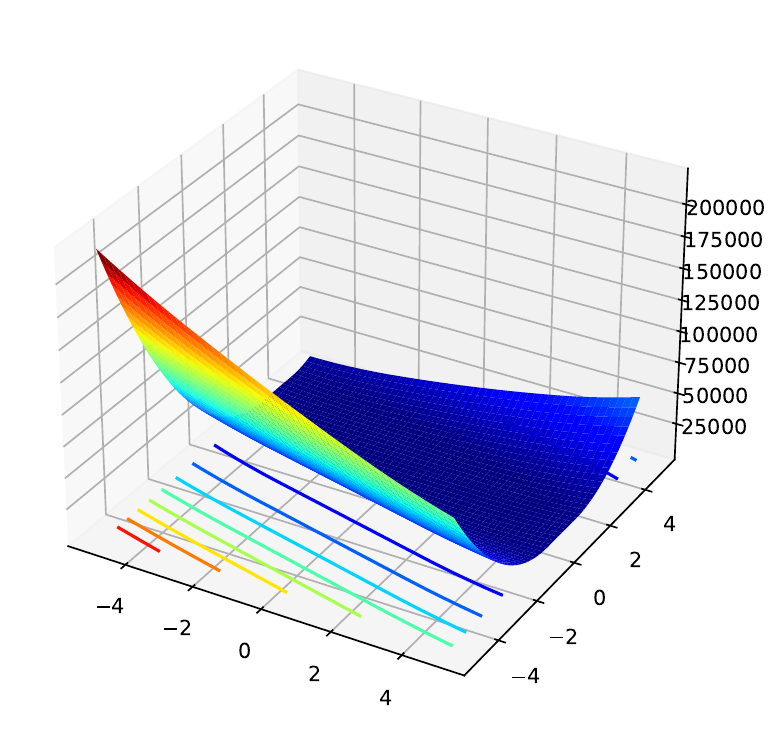}
    }
    \hfill
    \subfloat[F14: $f_{22}$, Gallagher’s Gaussian 21-hi Peaks]{
    \label{fig:func_22}
    \includegraphics[height=0.15\textwidth]{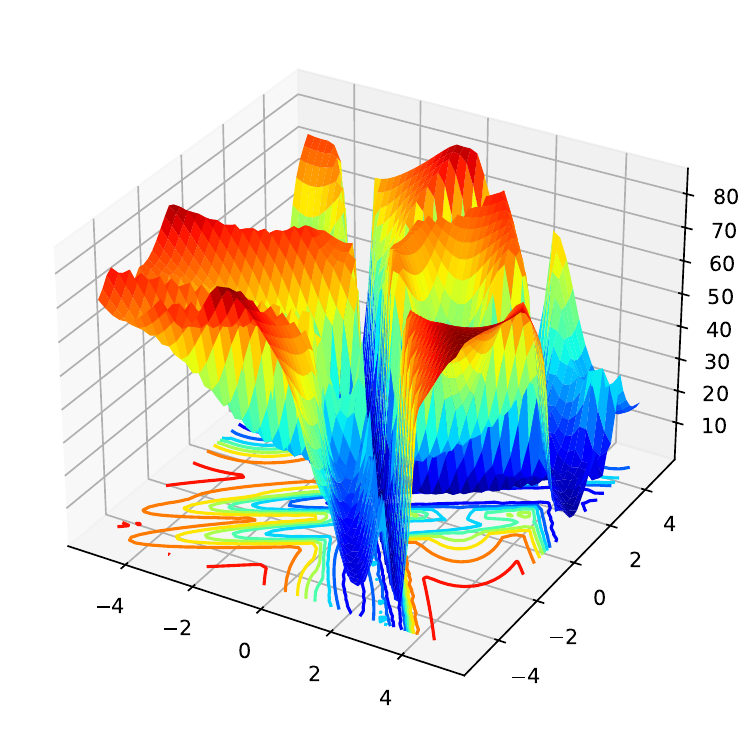}
    }
    \hfill
    \subfloat[F15: $f_{23}$, Katsuura]{
    \label{fig:func_23}
    \includegraphics[height=0.15\textwidth]{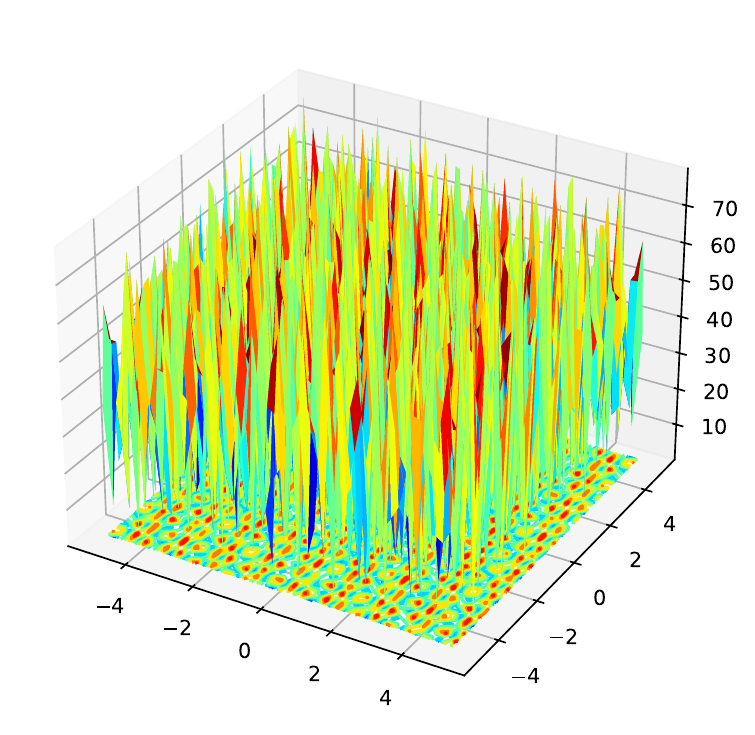}
    }
    \hfill
    \subfloat[F16: $f_{24}$, Lunacek bi-Rastrigin]{
    \label{fig:func_24}
    \includegraphics[height=0.15\textwidth]{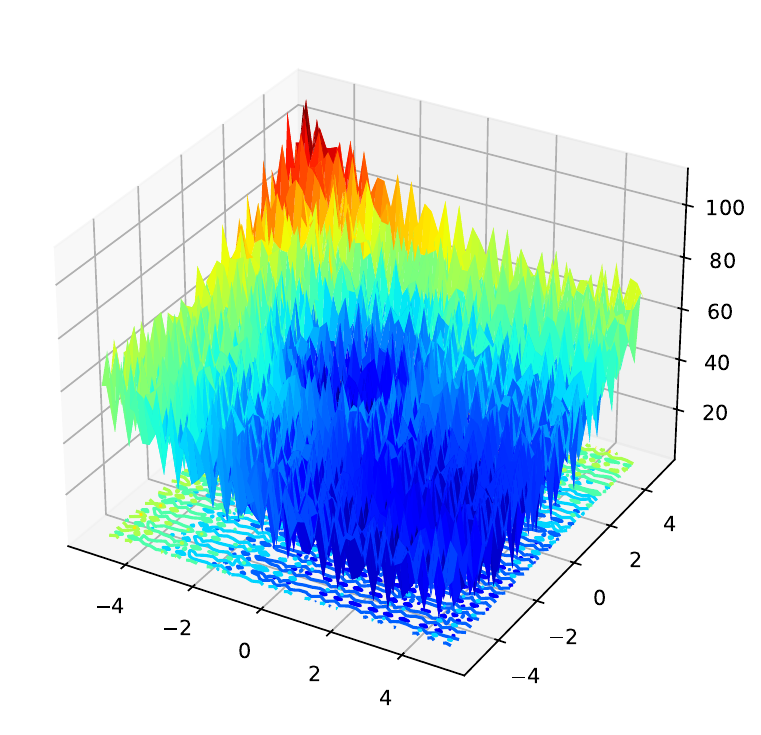}
    }
    \caption{Fitness landscapes of functions in the testing set when dimension is set to 2. }
    \label{fig:test_func}
\end{figure*}

MetaBox provides 24 synthetic problem instances with diverse characteristics and landscapes as listed in Table~\ref{tab:bbob}. Figure~\ref{fig:train_func} and Figure~\ref{fig:test_func} presents the landscapes of the training and testing problem instances when dimensions are set to 2, respectively. We select $f_1$, $f_2$, $f_3$, $f_5$, $f_{15}$, $f_{16}$, $f_{17}$ and $f_{21}$ as training functions and the rest for testing to balance the optimization difficulty in training and testing problem sets.

\begin{table*}[h]
    \centering
    \caption{Overview of the BBOB testsuites.}
    \label{tab:bbob}
    \resizebox{0.5\textwidth}{!}{
    \begin{tabular}{|p{3cm}<{\centering}|c|l|}
        \hline
        & Problem & Functions \\
        \hline
        \multirow{5}{*}{Separable functions}
        & $f_1$ & Sphere Function \\  \cline{2-3} 
        & $f_2$ & Ellipsoidal Function \\ \cline{2-3} 
        & $f_3$ & Rastrigin Function \\ \cline{2-3} 
        & $f_4$ & Buche-Rastrigin Function \\ \cline{2-3} 
        & $f_5$ & Linear Slope \\ \hline
        \multirow{4}{*}{\makecell{Functions \\ with low or moderate \\ conditioning}}
        & $f_6$ & Attractive Sector Function \\ \cline{2-3} 
        & $f_7$ & Step Ellipsoidal Function \\ \cline{2-3} 
        & $f_8$ & Rosenbrock Function, original \\ \cline{2-3} 
        & $f_9$ & Rosenbrock Function, rotated \\ \hline
        \multirow{5}{*}{\makecell{Functions with \\ high conditioning \\ and unimodal}}
        & $f_{10}$ & Ellipsoidal Function \\ \cline{2-3}
        & $f_{11}$ & Discus Function \\ \cline{2-3}
        & $f_{12}$ & Bent Cigar Function \\ \cline{2-3}
        & $f_{13}$ & Sharp Ridge Function \\ \cline{2-3}
        & $f_{14}$ & Different Powers Function \\ \hline
        \multirow{5}{*}{\makecell{Multi-modal \\ functions \\ with adequate \\ global structure}}
        & $f_{15}$ & Rastrigin Function (non-separable counterpart of F3) \\ \cline{2-3}
        & $f_{16}$ & Weierstrass Function \\ \cline{2-3}
        & $f_{17}$ & Schaffers F7 Function \\ \cline{2-3}
        & $f_{18}$ & Schaffers F7 Function, moderately ill-conditioned \\ \cline{2-3}
        & $f_{19}$ & Composite Griewank-Rosenbrock Function F8F2 \\ \hline
        \multirow{5}{*}{\makecell{Multi-modal \\ functions \\ with weak \\ global structure}}
        & $f_{20}$ & Schwefel Function \\ \cline{2-3}
        & $f_{21}$ & Gallagher’s Gaussian 101-me Peaks Function \\ \cline{2-3}
        & $f_{22}$ & Gallagher’s Gaussian 21-hi Peaks Function \\ \cline{2-3}
        & $f_{23}$ & Katsuura Function \\ \cline{2-3}
        & $f_{24}$ & Lunacek bi-Rastrigin Function \\ \hline
        \multicolumn{3}{|c|}{Default search range: [-5, 5]$^{D}$} \\ \hline
    \end{tabular}}
\end{table*}

\section{AEI Metric}

In Section 5.3.3 in the main paper, we present the best objective value AEI scores of the baselines for validating the zero-shot performance on realistic problems. 
To get the score, we test the target approach on $K$ problem instances for $G$ repeated runs and then record the basic metrics. The best objective value $v_{obj}^{k,g}$ denotes the best objective found by the algorithm on the $k$-th problem during the $g$-th test run. To convert its monotonicity, we conduct an inverse transformation, i.e., $v_{obj}^{k,g} = \frac{1}{v_{obj}^{k,g}}$.
Then, Z-score normalization is applied: 
\begin{equation}
    Z_{*}^{k} = \frac{1}{G}\sum_{g=1}^{G}\frac{v_{obj}^{k,g}-\mu_*}{\sigma_*},
\end{equation}
where $\mu_*$ and $\sigma_*$ are calculated by using Random Search as a baseline. Finally, the best objective value AEI score is calculated by: 
\begin{equation}
    AEI=\frac{1}{K}\sum_{k=1}^{K}e^{Z_{obj}^k},
\end{equation}
where Z-scores are first aggregated, then subjected to an inverse logarithmic transformation, and subsequently averaged across the test problem instances. A higher AEI indicates better performance of the corresponding approach.

\end{document}